\documentclass{article}

 \usepackage[main, final]{neurips_2025}
\usepackage[strings]{underscore}

\usepackage[utf8]{inputenc} 
\usepackage[T1]{fontenc}    
\usepackage{hyperref}       
\usepackage{url}            
\usepackage{booktabs}       
\usepackage{amsfonts}       
\usepackage{nicefrac}       
\usepackage{microtype}      
\usepackage{xcolor}         

\usepackage{graphicx}
\usepackage{multicol}
\usepackage{CJKutf8}
\newcommand{\chinese}[1]{ \begin{CJK*}{UTF8}{gbsn}#1\end{CJK*}}
\usepackage{makecell}

\title{MORQA: Benchmarking Evaluation Metrics \\for Medical Open-Ended Question Answering}

\author{%
  Wen-wai Yim\\
  Microsoft Health AI\\
  \texttt{yimwenwai@microsoft.com} \\
  \And
  Asma Ben Abacha\\
  Microsoft Health AI\\
  \texttt{abenabacha@microsoft.com} \\
  \And
  Zixuan Yu\\
  University of Washington\\
  \texttt{zixuanyu@uw.edu} \\
  \AND
  Robert Doerning\\
  University of Washington\\
  \texttt{doerning@uw.edu} \\
  \And
  Fei Xia\\
  University of Washington\\
  \texttt{fxia@uw.edu} \\
  \And
  Meliha Yetisgen\\
  University of Washington\\
  \texttt{melihay@uw.edu}
}

\begin{document}

\maketitle

\begin{abstract}
Evaluating natural language generation (NLG) systems in the medical domain presents unique challenges due to the critical demands for accuracy, relevance, and domain-specific expertise. Traditional automatic evaluation metrics, such as BLEU, ROUGE, and BERTScore, often fall short in distinguishing between high-quality outputs, especially given the open-ended nature of medical question answering (QA) tasks where multiple valid responses may exist.
In this work, we introduce MORQA (Medical Open-Response QA), a new multilingual benchmark designed to assess the effectiveness of NLG evaluation metrics across three medical visual and text-based QA datasets in English and Chinese. Unlike prior resources, our datasets feature 2-4+ gold-standard answers authored by medical professionals, along with expert human ratings for three English and Chinese subsets. We benchmark both traditional metrics and large language model (LLM)-based evaluators, such as GPT-4 and Gemini, finding that LLM-based approaches significantly outperform traditional metrics in correlating with expert judgments. We further analyze factors driving this improvement, including LLMs' sensitivity to semantic nuances and robustness to variability among reference answers. Our results provide the first comprehensive, multilingual qualitative study of NLG evaluation in the medical domain, highlighting the need for human-aligned evaluation methods. All datasets and annotations will be publicly released to support future research. 
\end{abstract}

\section{Introduction}

The rapid advancement of large language models (LLMs) and their emergent capabilities has enabled these models to generate high-quality answers in response to natural language questions. Automatic evaluation metrics are designed to compare system outputs against gold-standard references and provide a cost-effective means to assess output/answer quality. However, the limitations of traditional natural language generation (NLG) metrics, such as BLEU and ROUGE, are increasingly evident, particularly in open-ended generation tasks where multiple correct answers may exist.

Evaluation is especially challenging in the medical domain, where precision, contextual understanding, and expert reasoning are essential, and where multiple correct answers may vary significantly in both content and phrasing, while remaining equally valid. Traditional metrics often fail to capture this variability, resulting in unreliable assessments. At the same time, these metrics may assign high scores to responses that are lexically similar to the gold standard but factually incorrect or clinically misleading, further underscoring their limitations in sensitive, knowledge-intensive domains.
In this work, we present the first qualitative evaluation of state-of-the-art NLG evaluation metrics for open-ended medical question answering (QA). We benchmark performance across three datasets in two languages: English and Chinese. Unlike prior resources that rely on synthetic data or single-reference annotations, our datasets include 2–4+ gold-standard answers authored by certified medical professionals. Additionally, human evaluation is conducted by at least one domain expert per instance, with English clinical subsets assessed by two practicing medical doctors.

\begin{table} 
\centering 
\scalebox{.8}{
\begin{tabular}{|p{16cm}|} 
\hline
DermaVQA-iiyi \\
QUERY\_TITLE: \chinese{帮忙诊断一下} // Please help diagnose\\
QUERY\_CONTENT:
\chinese{三个月前出现如下图，自己用达克能宁喷雾两个月} \chinese{无明显效果，之后去乡村诊所，医生指导用鸡眼膏，之后出现变 红变多，请帮忙诊断下} // Three months ago, the condition shown in the picture below appeared. The patient used Daknening spray for two months without any noticeable effect. Afterwards, they went to a rural clinic, where the doctor advised them to use corn ointment. Subsequently, the condition turned red and worsened. Please help with the diagnosis.\\
\centering \makecell{ \includegraphics[width=30mm, height=30mm]{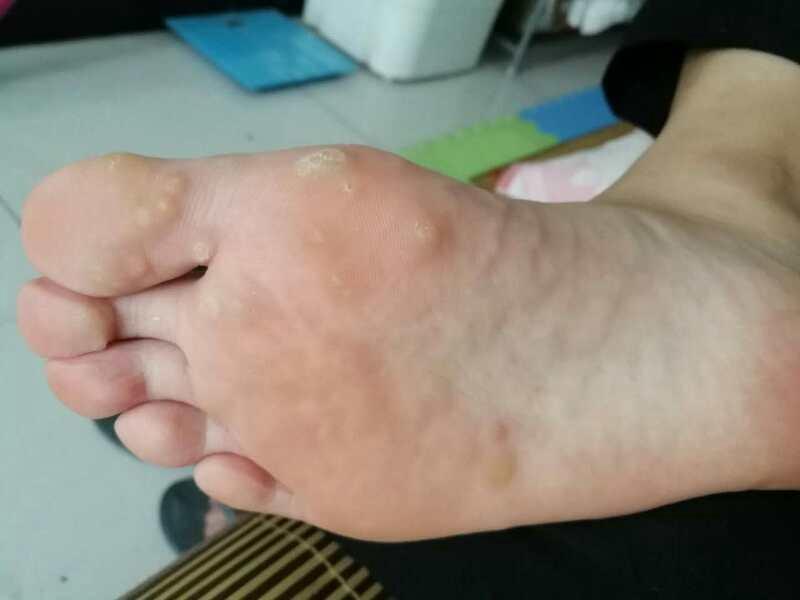}  \includegraphics[width=30mm, height=30mm]{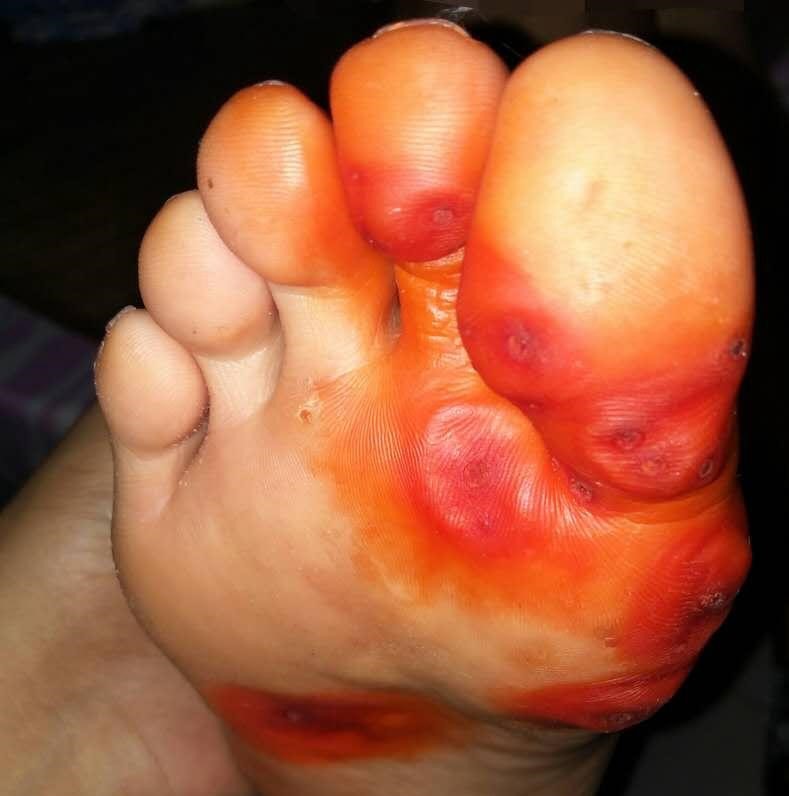} } \\
\raggedright RESPONSE: \chinese{考虑：跖疣。} // Consider Plantar wart.\\\hrule
WoundcareVQA  \\
QUERY\_TITLE: \chinese{求解啊 这怎么弄啊} // Seeking for the advice! What should I do? \\
QUERY\_CONTENT: \\
\chinese{一个半月了，不消肿 ，是摔得，伤口特别深，有半厘米左右，伤口的周围红肿，一按有小坑，而且半天不回去，跟水肿了似得，怎么办啊？} // It's been a month and a half, and the swelling hasn't gone down. It was a fall, and the wound was particularly deep, about half a centimeter. The area around the wound is red and swollen. When pressed, it forms a small pit that takes a long time to go back, looking like edema. What should I do? \\
\centering \makecell[b]{ \includegraphics[width=30mm, height=30mm]{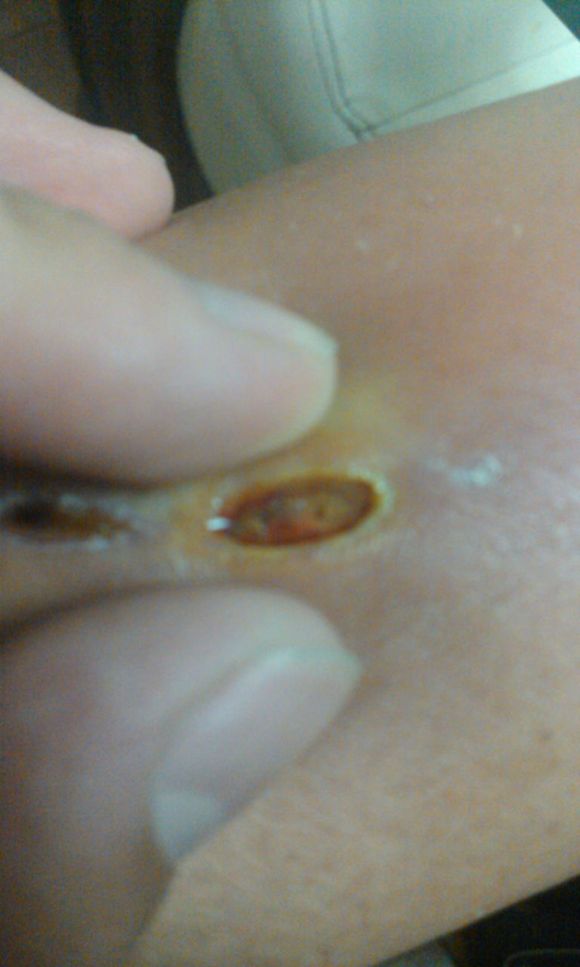}  \includegraphics[width=30mm, height=30mm]{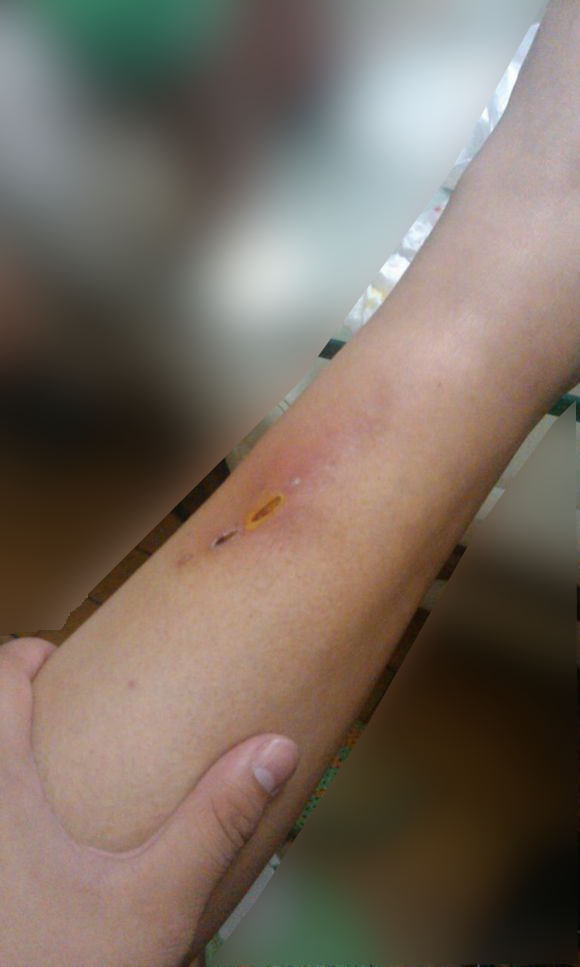} \includegraphics[width=30mm, height=30mm]{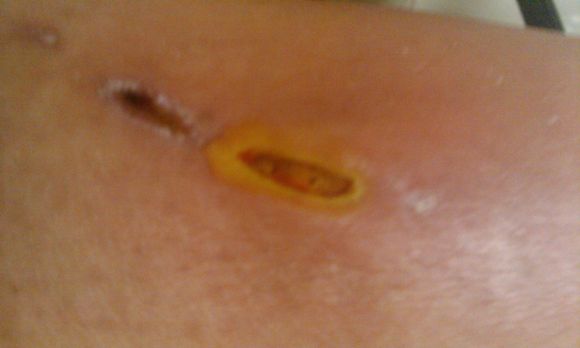} } \\
\raggedright RESPONSE: \chinese{伤口迟迟未能愈合、伤口周边有红肿，很可能是感染。建议到紧急护理诊所拿取抗生素服用。} //
Delayed wound healing and surrounding redness and swelling of the area around the wound is suspicious for infection. Recommend evaluation in urgent care for antibiotics.\\\hrule
LiveQA-clinical \\
QUERY\_TITLE: Ear Wax. \\
QUERY\_CONTENT: I sometimes drop Peroxide into the ear and let it bubble for a couple of minutes, then use warm water to flush it out. is there harm?\\
RESPONSE: To clean the ears, wash the external ear with a cloth, but do not insert anything into the ear canal. Most cases of ear wax blockage respond to home treatments used to soften wax. Patients can try placing a few drops of mineral oil, baby oil, glycerin, or commercial drops in the ear. Detergent drops such as hydrogen peroxide or carbamide peroxide (available in most pharmacies) may also aid in the removal of wax. Irrigation or ear syringing is commonly used for cleaning and can be performed by a physician or at home using a commercially available irrigation kit. Common solutions used for syringing include water and saline, which should be warmed to body temperature to prevent dizziness. Ear syringing is most effective when water, saline, or wax dissolving drops are put in the ear canal 15 to 30 minutes before treatment. \\\hrule
MedDialog-ZH  \\
QUERY\_CONTENT: \chinese{嘴唇皮脂腺异位症。嘴唇皮脂腺异位症连成一片白色影响美观和心里。}\\
RESPONSE: \chinese{可以高频电离子治疗。} \\\hrule
\end{tabular}
}
\caption{(V)QA Examples from the MORQA dataset sourced from the DermaVQA-iiyi, WoundcareVQA, LiveQA-Clinical \& MedDialog-ZH collections} 
\label{tab:dataset-examples}
\end{table}

\noindent Our key contributions are as follows:
\begin{itemize}
    \item We introduce MORQA\footnote{Code and data will be released upon publication.}, a new expert-annotated dataset of open-ended medical QA responses, spanning three English and three Chinese collections. Each query includes three or more system-generated responses and at least one expert rating per response, totaling 16,041 expert judgments.
    \item We benchmark current state-of-the-art NLG evaluation metrics for medical open-ended question answering across both text-based and visual question answering datasets. We also study these metrics when multiple gold responses are measured against each other.
    \item We analyze the capabilities of large language models as evaluators (LLM-as-a-judge) across various settings, including single vs. multiple references, referenceless evaluation, and multimodal inputs (with and without image support). 
\end{itemize}


Table~\ref{tab:dataset-examples} presents (visual) QA examples from the MORQA dataset, sourced from the DermaVQA-iiyi, WoundcareVQA, LiveQA-Clinical, and MedDialog-ZH collections. 

\section{Background}

Open-response question answering (QA) is a variant of QA in which the output is a free-form text answer, rather than a categorical label such as a multiple-choice option or a True/False decision. This task intersects with several related areas, including community QA, consumer health QA, medical dialogue generation, and visual question answering. In community QA (CQA), users post questions on public forums, and other users provide textual answers. These datasets span a wide range of domains, from general-purpose platforms like Tieba~\cite{Tian2017HowTM}, to more specialized forums focused on online shopping (JDC) and programming (Ubuntu)~\cite{Zhang2019ReCoSaDT}. Other sources include Stack Overflow and Yahoo! Answers~\cite{Srba2016ACS}.


In the medical domain, consumer health QA (CHQA) involves answering health-related questions, often submitted by users to public health information websites such as those maintained by the U.S. National Library of Medicine~\cite{LiveMedQA2017}. Medical dialogue generation tasks further expand this to include multi-turn interactions between patients and healthcare providers~\cite{zeng-etal-2020-meddialog,Liu2020MedDGAE}. 


Visual Question Answering (VQA) tasks involve visual inputs, such as images or diagrams, accompanied by a user query. While most VQA research has focused on categorical or short-form textual answers (e.g., multiple choice or one- to two-word responses), open-response VQA remains relatively underexplored. In the general domain, VQAOnline~\cite{vqaonline} presents a dataset of Stack Exchange questions paired with relevant images. In the medical domain, VQARad~\cite{lau2018dataset} includes a subset of longer free-text responses. More recently, DermaVQA~\cite{Yim2024DermaVQAAM} and WoundCareVQA~\cite{YIM2025104888} have introduced datasets composed entirely of sentence- or paragraph-level answers.

Across these tasks, responses are typically evaluated using a range of natural language generation metrics, including BLEU, ROUGE, and METEOR. However, the specific set of metrics varies by subdomain. For instance, VQAOnline additionally reports CLIP-S scores to capture image-text similarity. For reference, we summarize performance scores across tasks and datasets in Table~\ref{tab:qadatasets}. Of the datasets, only two (WoundcareVQA and VQAOnline) have comparisons with human evaluation. As shown, evaluation results are both highly dependent on the chosen metric and highly variable across datasets. For example, BLEU scores range from 0.007 to 0.123 for the MedDialog dataset, a BLEU score of 0.018 corresponds to a METEOR score of 0.131. Similarly, for WoundCareVQA, a ROUGE-L score of 0.501 co-occurs with a BLEU score of just 0.123. This variability underscores the challenge of evaluating open-response QA and highlights the importance of studying medical open-response QA across multiple datasets and metrics. 

\begin{table}
  \centering
  \scalebox{0.75}{
  \begin{tabular}{cccc}
    \hline
    \textbf{Dataset} & \textbf{Domain} & \textbf{\#Ref} & \textbf{Scores}\\
    \hline
    \underline{Community QA} & &\\
    JDC & e-commerce & 1 & BLEU: 0.138 \\
    Ubuntu Dialog & coding & 1 & BLEU: 0.0017\\
    Tieba & general & 1 & BLEU: 0.007\\
    \hline
    \multicolumn{2}{l}{\underline{Consumer Health QA} } &&\\
    LiveQA* & health & 1-2 & --\\
    \hline
    \multicolumn{2}{l}{\underline{Medical Dialogue Generation} } &&\\
    MedDialog & medical & 1 & BLEU: 0.018\\
    &  &  & METEOR: 0.131\\
    MedDG & medical & 1 & BLEU: 0.240 \\
    \hline
    \underline{VQA} & &&\\
    DermaVQA &  medical & 2+ & BERTSCORE: 0.856\\
    &   &  & BLEU:  0.093\\
    WoundcareVQA &  medical & 3 & BERTSCORE: 0.692\\
    &  &  & BLEU: 0.123\\
    &  &  & ROUGE-L: 0.501\\
    VQAOnline &  general & 1 & ROUGE-L: 0.15\\
     &   &  & METEOR: 0.14\\
    VQARad\textdagger &  medical & 1 & BLEU: 0.1047\\
    \hline
  \end{tabular}
  }
  \caption{QA Evaluation performances for Human Generated Datasets. (\#Ref = Number of references in the test set, *LiveQA did not have automatic ratings, \textdagger VQARad subset described here includes only open-ended instances)}
  \label{tab:qadatasets}
\end{table}

\section{Related Works}

Human evaluation remains the gold standard for assessing NLG systems, but it is time-consuming and costly. Beginning with machine translation efforts in the 1950s, evaluation has played a central role in the text generation field. This remains true through the development of transformer-based generation models and into the current era of LLMs. As NLG systems have evolved, the demand for effective evaluation methodologies has grown accordingly and now spans across multiple tasks such as machine translation, summarization, data-to-text generation, image captioning, dialogue generation, and open-domain question answering. 

\textbf{Automatic Metrics.} Prior to the development of efficient neural network training and the popularization of dense distributional word embeddings, automatic evaluation metrics mainly relied on word-based ngram metrics such as the precision-based BLEU~\cite{Papineni2002BleuAM} and the Recall-based ROUGE~\cite{Lin2004ROUGEAP}. Starting in 2014, new metrics utilizing embeddings such as BERTScore~\cite{Zhang2019BERTScoreET}, MoverScore~\cite{Zhao2019MoverScoreTG}, BLEURT~\cite{Sellam2020BLEURTLR} became available. More recently, following the success of large transformer models such as GPT-3, scoring methods that leverage likelihoods computed by pretrained language models, such as BARTScore~\cite{Yuan2021BARTScoreEG} and GPTScore~\cite{Fu2023GPTScoreEA}, have  been proposed. Recently, with powerful LLMs, direct assessment (LLM-as-a-judge) have also been proposed and shown superior performance~\cite{Zheng2023JudgingLW}. Extending this work, there is now active area of research related to fine-tuning LLM for better judgment capabilities by taking instructions/rubrics inputs~\cite{Kim2024Prometheus2A,Ye2023FLASKFL,gupta2025carmodynamiccriteriageneration}.

\textbf{Human Metrics.} Human evaluation is typically captured as a rating on a fixed numerical scale with attached quality levels. In QA datasets, several studies have incorporated human assessments. For instance, in the LiveQA dataset, part of the TREC 2017 challenges, a 1-4 rating scale was used, with 4 indicating the highest quality~\cite{LiveMedQA2017}. In the VQAOnline Stack Exchange and WoundcareVQA datasets, raters applied scores of {0, 0.5, 1}, corresponding to incorrect, partially correct, and correct answers, respectively~\cite{vqaonline,YIM2025104888}. More recently, large-scale evaluation datasets have emerged, either synthetically generated, such as the Feedback Collection~\cite{Kim2023PrometheusIF}, or gathered through platforms like ChatbotArena~\cite{Zheng2023JudgingLW}, using 1-5 rating scales and/or pairwise comparison judgments. 

\section{Evaluation Methodology}

\subsection{Datasets}

\textbf{Visual Question Answering:} WoundcareVQA~\cite{YIM2025104888} and DermaVQA~\cite{Yim2024DermaVQAAM} are multi-modal open-response QA datasets, where each patient query and 1+ images are paired with 2+ free text responses from medical doctors. English and Chinese are versions available. These datasets were selected as they included naturally occurring medical queries. As the original DermaVQA responses were sourced from online forums, we extend the dataset by adding two additional high-quality responses from board-certified dermatologists per instance, which are included as reference answers. 

\noindent \textbf{Textual Question Answering:} In addition to the multi-modal open-response VQA datasets, we include two textual QA datasets: LiveQA~\cite{LiveMedQA2017} 
and MEDDIALOG~\cite{zeng-etal-2020-meddialog}. LiveQA, originally part of the TREC 2017 challenge, includes consumer health questions posed to US National Institutes of Health sites, answered by domain experts. We filtered the LiveQA test set to include patient-related questions and excluded other cases (e.g., asking for open hours, or references to literature). The MedDialog dataset includes 2-speaker multi-turn online patient-doctor conversations from icliniq.com, healthcaremagic.com, and haodf.com, encompassing a variety of medical specialties. We specifically used the chunyudoctor subset and filtered for 2-turn conversations (one patient query, one doctor response) and randomly sampled 500 cases. Each case was further manually reviewed and excluded if the question or answer was irrelevant or incomplete. The final statistics for the (visual) QA pairs included in the \textsc{MORQA} dataset are presented in Table~\ref{tab:datastatistics}.

\begin{table}
  \centering
  \scalebox{0.7}{
  \begin{tabular}{ccccccc}
    \hline
     & \textbf{\#Query} & \multicolumn{2}{c}{\textbf{\#Responses} }  & \textbf{Lang} & \textbf{\#Raters} & \textbf{\#Hum-evals} \\ 
     & & \textbf{Gold} & \textbf{Sys} &  & \textbf{EN/ZH} & \textbf{EN/ZH}\\ \hline
    \multicolumn{6}{l}{\underline{WoundcareVQA}} & \\ 
    valid & 105 & 210 & 315 & \{en,zh\} & 1/1& 6/2\\ 
    test & 93 & 279 & 279 & \{en,zh\} & 2/1 & 6/2\\ 
    \multicolumn{6}{l}{\underline{DermaVQA}}  & \\ 
    valid & 56 & 417 & 158 & \{en,zh\} & 2/1 & 6/2\\ 
    test & 100 & 926 & 300 & \{en,zh\} & 2/1 & 6/2\\ 
    \multicolumn{6}{l}{\underline{LiveQA}}  & \\ 
    test & 40 & 62 & 233 & \{en\} & 1 & 1\\ 
    \multicolumn{6}{l}{\underline{MedDialog}}  & \\ 
    valid & 236 & 236 & 708 & \{zh\} & 1& 2\\ 
    test & 259 & 259 & 777 & \{zh\} & 1 & 2\\  \hline
  \end{tabular}}
  \caption{Statistics of the \textsc{MORQA} dataset after filtering and annotation, including the number of (visual) questions, system and gold answers, and human evaluation ratings.} 
  \label{tab:datastatistics}
\end{table}

\subsection{Human Evaluation Ratings}

To conduct this study, we (i) generated system responses and (ii) collected domain expert evaluations for each response. For the DermaVQA, WoundcareVQA, and MedDialog datasets, responses were generated using GPT-4o, Gemini-1.5-pro-002, and Qwen-VL-Chat. These were chosen for their multi-modal capabilities as well as their general capabilities in English and Chinese.

For human ratings, the DermaVQA English subsets were labeled by two  practicing dermatologists. The WoundCareVQA English subsets were labeled by two practicing medical doctors specializing in General Surgery and Emergency Medicine. They graded each response on a 3-point scale {0.0, 0.5, 1.0} where 1.0 is the best. The measured qualities include (1) disagreement, (2) completeness, (3) factual accuracy, (4) relevance, (5) writing style, and (6) overall quality. Gold references were available, however, raters were allowed to use their own independent judgment that may diverge from references.

For the Chinese datasets (DermaVQA, WoundCareVQA, and MedDialog), responses were labeled by one domain expert with Chinese medical school training (without residency training and not currently practicing). We adjusted the human ratings to measure consistency with gold references. A 3-point scale \{0.0, 0.5, 1.0\}, where 1.0 indicates the highest quality, was used to evaluate (1) factual consistency with the reference and (2) general appropriate writing style from doctor to patient. 

For LiveQA responses, we used the released system responses and human ratings provided by the TREC organizers. These ratings follow a 0-4 scale, where 4 denotes the highest quality. Summary statistics, including the total number of human evaluation ratings, are shown in Table \ref{tab:datastatistics}. 

\subsection{Automatic Evaluation Metrics}

We evaluated multiple automatic NLG evaluation metrics including classical n-gram-based metrics, embedding similarity-based metrics, learned model-based metrics, as well as instruction-following LLM-based direct evaluation. The specific metrics and corresponding references are detailed below:

\begin{itemize}
    \item \underline{n-gram-based Metrics}: BLEU~\cite{galley-etal-2015-deltableu}, ROUGE~\cite{lin-2004-rouge} 
    \item \underline{Embedding Similarity}: BERTScore~\cite{Zhang2019BERTScoreET}
    \item \underline{Learned Metrics}: BLEURT~\cite{Sellam2020BLEURTLR}
    \item \underline{LLM-based Evaluation}: Prometheus-v2-7b~\cite{Kim2024Prometheus2A}, GPT-4o\footnote{openai.com/index/hello-gpt-4o}, Gemini-1.5-pro-002\footnote{ai.google.dev/gemini-api/docs/models\#gemini-1.5-pro}, DeepSeekV3~\cite{DeepSeekAI2024DeepSeekV3TR}
\end{itemize} 

Model configurations and details are available in the Appendix section.

\subsection{Analysis}

Automatic evaluation metrics were measured against human evaluation ratings using common correlation methods: Spearman's rank correlation, Pearson's correlation, and Kendall's rank correlation. We also calculated pairwise ranking accuracies. Specifically, for each pair of systems, we evaluated whether the predicted win/loss/tie outcome matched the corresponding human judgment. For numerical metrics such as BERTScore, a tie was assigned when the absolute difference between scores was less than 0.05. 

\begin{table*} 
\centering 
\scalebox{.65}{ 
\begin{tabular}{c|ccccc|ccccc|ccccc} 
\hline 
& \multicolumn{5}{c}{\bf DermaVQA} & \multicolumn{5}{|c}{\bf LiveQA} & \multicolumn{5}{|c}{\bf WoundcareVQA} \\
\hline 
metric  & $\tau$ & $r$ & $\rho$ & avg-corr & acc & $\tau$ & $r$ & $\rho$ & avg-corr & acc & $\tau$ & $r$ & $\rho$ & avg-corr & acc\\ \hline
bertscore-f1-max & 0.29 & 0.33 & 0.37 & 0.33 & 0.57 & 0.38 & 0.39 & 0.48 & \underline{0.41} & 0.29 & 0.18 & 0.25 & 0.22 & 0.22 & 0.45 \\
bertscore-f1-mean & 0.30 & 0.36 & 0.39 & 0.35 & 0.57 & 0.36 & 0.38 & 0.46 & 0.40 & 0.29 & 0.16 & 0.24 & 0.20 & 0.20 & 0.46 \\
bleurt20-max & 0.33 & 0.35 & 0.42 & 0.37 & 0.55 & 0.07 & 0.08 & 0.09 & 0.08 & 0.19 & 0.22 & 0.30 & 0.27 & 0.26 & 0.41 \\
bleurt20-mean & 0.33 & 0.40 & 0.42 & \underline{0.38} & 0.54 & 0.08 & 0.08 & 0.10 & 0.09 & 0.20 & 0.16 & 0.25 & 0.20 & 0.20 & 0.43 \\
deepseekv3 & 0.32 & 0.34 & 0.35 & 0.34 & \underline{0.59} & 0.36 & 0.36 & 0.38 & 0.37 & 0.62 & 0.37 & 0.43 & 0.40 & \textbf{0.40} & \underline{0.60} \\
deltableu-micro & 0.23 & 0.17 & 0.29 & 0.23 & 0.51 & 0.17 & 0.23 & 0.22 & 0.21 & 0.27 & 0.16 & 0.17 & 0.20 & 0.18 & 0.42 \\
gpt-4o & 0.36 & 0.38 & 0.39 & \underline{0.38} & 0.58 & 0.38 & 0.39 & 0.40 & 0.39 & \textbf{0.65} & 0.36 & 0.41 & 0.38 & \underline{0.38} & 0.59 \\
gemini-1.5-pro & 0.40 & 0.42 & 0.42 & \textbf{0.41} & \textbf{0.62} & 0.43 & 0.40 & 0.46 & \textbf{0.43} & \underline{0.64} & 0.36 & 0.38 & 0.37 & 0.37 & \textbf{0.64} \\
promethesus7bv2 & 0.14 & 0.14 & 0.15 & 0.14 & 0.55 & 0.26 & 0.24 & 0.28 & 0.26 & 0.61 & 0.19 & 0.23 & 0.20 & 0.20 & 0.59 \\
rouge-1 & 0.04 & 0.10 & 0.05 & 0.06 & 0.46 & 0.13 & 0.11 & 0.16 & 0.13 & 0.25 & 0.04 & 0.21 & 0.05 & 0.10 & 0.39 \\
rouge-2 & 0.13 & 0.17 & 0.17 & 0.16 & 0.46 & 0.23 & 0.25 & 0.30 & 0.26 & 0.29 & 0.10 & 0.23 & 0.13 & 0.15 & 0.40 \\
rouge-L & 0.18 & 0.21 & 0.23 & 0.20 & 0.49 & 0.16 & 0.13 & 0.20 & 0.16 & 0.25 & 0.10 & 0.22 & 0.12 & 0.15 & 0.44 \\ \hline
\end{tabular}
}
\caption{Correlations with Human Overall Metric - English ($\tau$=kendall's tau, $r$=pearson, $\rho$=spearman, avg-corr=mean of $\tau$, $r$, $\rho$) }
\label{tab:results-en-overall}
\end{table*}

\begin{table*} 
\centering 
\scalebox{.65}{ 
\begin{tabular}{c|ccccc|ccccc|ccccc} 
\hline 
 & \multicolumn{5}{c}{\bf DermaVQA} & \multicolumn{5}{|c}{\bf Meddialog} & \multicolumn{5}{|c}{\bf WoundcareVQA} \\
\hline 
metric  & $\tau$ & $r$ & $\rho$ & avg-corr & acc & $\tau$ & $r$ & $\rho$ & avg-corr & acc & $\tau$ & $r$ & $\rho$ & avg-corr & acc\\ \hline
bertscore-f1-max & 0.34 & 0.41 & 0.42 & \textbf{0.39} & 0.26 & 0.31 & 0.38 & 0.39 & \textbf{0.36} & 0.23 & 0.10 & 0.13 & 0.12 & 0.12 & 0.28 \\
bertscore-f1-mean & 0.30 & 0.36 & 0.37 & 0.34 & 0.27 & 0.31 & 0.38 & 0.39 & \textbf{0.36} & 0.23 & 0.10 & 0.14 & 0.12 & 0.12 & 0.29 \\
bleurt20-max & 0.25 & 0.31 & 0.31 & 0.29 & 0.26 & 0.27 & 0.32 & 0.34 & \underline{0.31} & 0.19 & 0.18 & 0.19 & 0.23 & 0.20 & 0.26 \\
bleurt20-mean & 0.21 & 0.28 & 0.27 & 0.25 & 0.26 & 0.27 & 0.32 & 0.34 & \underline{0.31} & 0.19 & 0.19 & 0.22 & 0.24 & 0.22 & 0.25 \\
deepseekv3 & 0.36 & 0.39 & 0.39 & \underline{0.38} & \underline{0.55} & 0.29 & 0.30 & 0.30 & 0.29 & \underline{0.57} & 0.34 & 0.39 & 0.36 & \underline{0.36} & \textbf{0.54} \\
deltableu-micro & 0.28 & 0.30 & 0.35 & 0.31 & 0.26 & 0.24 & 0.25 & 0.30 & 0.26 & 0.21 & 0.10 & 0.11 & 0.13 & 0.11 & 0.28 \\
gpt-4o & 0.36 & 0.39 & 0.38 & \underline{0.38} & 0.54 & 0.28 & 0.30 & 0.29 & 0.29 & 0.53 & 0.38 & 0.41 & 0.39 & \textbf{0.39} & \textbf{0.54} \\
gemini-1.5-pro & 0.37 & 0.40 & 0.38 & \underline{0.38} & \textbf{0.62} & 0.30 & 0.30 & 0.31 & 0.30 & \textbf{0.58} & 0.29 & 0.33 & 0.30 & 0.31 & \underline{0.51} \\
promethesus7bv2 & 0.14 & 0.12 & 0.15 & 0.14 & 0.48 & 0.13 & 0.16 & 0.13 & 0.14 & 0.47 & 0.07 & 0.10 & 0.07 & 0.08 & 0.38 \\
rouge-1 & 0.14 & 0.19 & 0.18 & 0.17 & 0.25 & 0.15 & 0.18 & 0.19 & 0.17 & 0.20 & 0.05 & 0.07 & 0.06 & 0.06 & 0.25 \\
rouge-2 & 0.33 & 0.40 & 0.41 & \underline{0.38} & 0.27 & 0.25 & 0.31 & 0.31 & 0.29 & 0.22 & 0.13 & 0.16 & 0.17 & 0.15 & 0.28 \\
rouge-L & 0.09 & 0.15 & 0.12 & 0.12 & 0.21 & 0.13 & 0.16 & 0.17 & 0.15 & 0.19 & -0.01 & 0.02 & -0.01 & -0.00 & 0.24 \\
\hline
\end{tabular}}
\caption{Correlations with Human Factual Consistency Metric - Chinese ($\tau$=kendall's tau, $r$=pearson, $\rho$=spearman, avg-corr=mean of $\tau$, $r$, $\rho$) }
\label{tab:woundcare-zh-factual-consistency-wgold-1}
\end{table*}

\begin{table*} 
\centering 
\scalebox{.8}{ 
\begin{tabular}{c|ccccc|ccccc} 
\hline 
 & \multicolumn{5}{c}{\bf DermaVQA} & \multicolumn{5}{|c}{\bf WoundcareVQA} \\
\hline 
metric  & $\tau$ & $r$ & $\rho$ & avg-corr & acc & $\tau$ & $r$ & $\rho$ & avg-corr & acc \\ \hline
 \underline{\textbf{completeness}} & & & & & & & & & &  \\
deepseekv3 & 0.16 & 0.18 & 0.17 & 0.17 & \textbf{0.61} & 0.29 & 0.34 & 0.31 & \textbf{0.31} & 0.61 \\
gpt-4o & 0.22 & 0.24 & 0.23 & 0.23 & 0.59 & 0.27 & 0.31 & 0.28 & 0.29 & 0.61 \\
gemini-1.5-pro & 0.22 & 0.25 & 0.23 & 0.23 & 0.60 & 0.27 & 0.32 & 0.28 & 0.29 & \textbf{0.64} \\
promethesus7bv2 & 0.23 & 0.23 & 0.25 & \textbf{0.24} & 0.57 & 0.19 & 0.24 & 0.20 & 0.21 & 0.60 \\
 \underline{\textbf{factual-accuracy}} & & & & & & & & & &  \\
deepseekv3 & 0.32 & 0.32 & 0.35 & 0.33 & 0.60 & 0.32 & 0.38 & 0.34 & \textbf{0.35} & 0.58 \\
gpt-4o & 0.32 & 0.29 & 0.34 & 0.32 & 0.57 & 0.33 & 0.37 & 0.34 & \textbf{0.35} & 0.59 \\
gemini-1.5-pro & 0.39 & 0.38 & 0.41 & \textbf{0.39} & \textbf{0.61} & 0.31 & 0.34 & 0.31 & 0.32 & \textbf{0.61} \\
promethesus7bv2 & 0.10 & 0.09 & 0.11 & 0.10 & 0.52 & 0.11 & 0.15 & 0.12 & 0.13 & 0.57 \\
 \underline{\textbf{relevance}} & & & & & & & & & &  \\
deepseekv3 & 0.34 & 0.34 & 0.36 & 0.35 & 0.59 & 0.20 & 0.24 & 0.21 & 0.21 & 0.57 \\
gpt-4o & 0.34 & 0.31 & 0.36 & 0.34 & \textbf{0.61} & 0.24 & 0.29 & 0.25 & \textbf{0.26} & \textbf{0.65} \\
gemini-1.5-pro & 0.38 & 0.39 & 0.41 & \textbf{0.39} & \textbf{0.61} & 0.22 & 0.26 & 0.22 & 0.24 & 0.63 \\
promethesus7bv2 & 0.07 & 0.04 & 0.07 & 0.06 & 0.50 & 0.22 & 0.25 & 0.22 & 0.23 & 0.60 \\
 \underline{\textbf{writing-style}} & & & & & & & & & &  \\
deepseekv3 & 0.23 & 0.27 & 0.25 & 0.25 & 0.64 & 0.17 & 0.21 & 0.18 & 0.19 & 0.57 \\
gpt-4o & 0.26 & 0.30 & 0.28 & 0.28 & 0.60 & 0.21 & 0.25 & 0.21 & 0.22 & \textbf{0.64} \\
gemini-1.5-pro & 0.27 & 0.31 & 0.28 & \textbf{0.29} & \textbf{0.65} & 0.22 & 0.25 & 0.22 & \textbf{0.23} & 0.61 \\
promethesus7bv2 & 0.14 & 0.12 & 0.15 & 0.14 & 0.56 & 0.17 & 0.22 & 0.18 & 0.19 & 0.60 \\
\hline
\end{tabular}}
\caption{DermaVQA \& WoundcareVQA - English ($\tau$=kendall's tau, $r$=pearson, $\rho$=spearman, avg-corr=mean of $\tau$, $r$, $\rho$) }
\label{tab:DermaVQA/WoundcareVQA-en}
\end{table*}

\begin{table*}
  \centering
  \scalebox{.9}{
  \begin{tabular}{c|cccc|cccc}
    \hline
    \bf Dataset & \multicolumn{4}{c}{\bf GPT-4o} &  \multicolumn{4}{|c}{\bf Gemini-1.5-pro-002} \\  \hline
     \it Correlation & 3 refs & 2 refs & 1 ref & 0 ref  & 3 refs & 2 refs & 1 ref & 0 ref \\  \hline
      \textbf{DermaVQA-iiyi} &  & & & & & &  \\ 
\it Kendall  & 0.363 & 0.372 & \bf 0.404 & 0.171  & 0.397 & \bf 0.439 & 0.423 & 0.124 \\
\it Pearson & 0.384 & 0.396 & \bf 0.427 & 0.194 & 0.422 & \bf 0.467 & 0.452 & 0.156 \\
\it Spearman & 0.392 & 0.403 & \bf 0.436 & 0.188  & 0.423 & \bf 0.468 & 0.453 & 0.134 \\
\it Average  & 0.380 & 0.390 & \bf 0.422 & 0.184  & 0.414 & \bf 0.458 & 0.443 & 0.138 \\   \hline
     \textbf{WoundcareVQA} & & & & & & & \\
\it Kendall  & \bf 0.365 & 0.331 & 0.295 & 0.211 & \bf 0.360 & 0.277 & 0.231 & 0.182 \\
\it Pearson  & \bf 0.406 & 0.386 & 0.344 & 0.296 & \bf 0.380 & 0.316 & 0.272 & 0.209 \\
\it Spearman & \bf 0.380 & 0.345 & 0.308 & 0.220  & \bf 0.372 & 0.286 & 0.239 & 0.188 \\
\it Average  & \bf 0.383 & 0.354 & 0.316 & 0.242 & \bf 0.371 & 0.293 & 0.248 & 0.193 \\
  \hline
  \end{tabular}
  }
  \caption{Ablation results showing the effect of varying the number of gold-standard reference answers (3, 2, 1, 0 ref) on the performance of LLM-based evaluators. In the zero-reference condition, evaluation was based on the input query for text-based QA dataset (LiveQA) and on both the query and input images for the visual QA datasets (DermaVQA-iiyi and WoundcareVQA). Average correlations are reported.} 
  \label{tab:ablation}
\end{table*}

\section{Results}

Table \ref{tab:results-en-overall} presents the correlation scores and pairwise ranking accuracies for the English datasets. DeepSeekV3, GPT-4o, and Gemini-1.5-pro demonstrated strong performance across all three datasets. The highest performances were by one of the three models, with second-place performances often by one of the other two models. 

Table \ref{tab:woundcare-zh-factual-consistency-wgold-1} presents the factual consistency results for the Chinese datasets. Compared to the English evaluations, the three LLM API models had strong correlations and accuracies.  Interestingly, in contrast to the English results, traditional metrics such as BERTScore and BLEURT showed relatively strong correlations with human judgments on the DermaVQA and MedDialog datasets. 

To examine whether correlation strengths are consistent across different human evaluation criteria, Table \ref{tab:DermaVQA/WoundcareVQA-en} shows correlations and accuracies for the other human evaluation axes in the DermaVQA and WoundcareVQA datasets. While factual accuracy and relevance have similar magnitudes of correlations and accuracies with the overall metric, the other two criteria, completeness and writing style, had a less prominent similarity. 

In summary, LLM-as-a-judge methods showed strong performances across datasets and languages. However, more modest metrics like BERTScore and BLEURT performed competitively on the Chinese subsets DermaVQA and MedDialog, likely due to the shorter sentence lengths in these responses. We also observed differences in performance across specific evaluation criteria: while completeness and writing style showed weaker alignment, relevance and factual accuracy, which likely contribute more heavily to the overall score, maintained strong correlations and accuracies.


\section{Ablation Study: Impact of Reference Grounding on LLM-based Evaluation}

We conducted an ablation study to assess the impact of varying the number of gold-standard references on the performance of LLM-based evaluators. Specifically, we evaluated model outputs using three, two, one, and zero expert references, progressively reducing the level of reference-based grounding. In the reference-free setting, evaluators relied on both the query and associated images for visual QA datasets. This setup enabled us to isolate the contribution (and potential bias) introduced by reference answers, and to evaluate the robustness of LLM-based evaluators under varying degrees of contextual grounding. 

Table~\ref{tab:ablation} presents the results of this ablation study. Increasing the number of references from 1 to more than 1 generally led to a 5–10 percentage point improvement in average correlation with human ratings. In the case of DermaVQA, however, adding more than one reference did not consistently improve correlations. This may be due to the variability and randomness of the available reference responses, which can include up to several different answers in some cases. 

\begin{table}
  \centering
  \scalebox{1.0}{
  \begin{tabular}{ccc}
    \hline
      & \underline{Gold-Holdout} & \underline{GPT-4o}\\
     \hline 
     \underline{\textbf{EN}} & \\
     BLEU  & 0.0469 &  \bf 0.0622\\
     BERTScore  & 0.6634 & \bf 0.6640\\
     GPT-4o-as-judge  & 0.5520  & \bf 0.6147\\
     ROUGE-L  & 0.4353 & \bf 0.4499\\
    \hline  
    Human-Expert  &  \bf 0.975 & 0.8629 \\
     \hline 
     \underline{\textbf{ZH}} & \\
     BLEU  & 0.1109 & \bf 0.1234\\
     BERTScore  & 0.6896 & \bf 0.6923 \\
     GPT-4o-as-judge  & 0.6595 & \bf 0.7832 \\
     ROUGE-L  & 0.2254 &  \bf 0.4961 \\ \hline 
     Human-Expert  &  \bf 0.975 & 0.6613 \\
     \hline 
  \end{tabular}}
  \caption{Evaluation results for the WoundCareVQA Holdout-One-Gold experiment. Each row compares (i) a gold response evaluated against two other gold references (\textit{Gold-Holdout}) and (ii) a GPT-4o response evaluated against all references.}
  \label{tab:hold-out-experiment}
\end{table}

\section{Automatic Metric Scores when References Are Measured Against References}

In classification tasks, inter-annotator agreement provides an upper bound on task performance, highlighting the inherent limit when even domain experts disagree. In NLG, two additional confounding factors arise: (a) measurement uncertainty and (b) variations in the gold standards, both in content and surface form. To study the uncertainty related to this, we leverage the multiple gold-standard responses available in the WoundcareVQA datasets, holding one gold standard response as a candidate system response, and compare it against the remaining two gold references. For comparison, we present GPT4-o generated responses compared against all three gold references. Evaluation was conducted by calculating several automatic metrics, including n-gram, embedding-based, and LLM-as-a-judge models. For human evaluation, we define a human expert score as the proportion of other gold responses that a human annotator judged as acceptable (excluding their own authored response).


The results, presented in Table \ref{tab:hold-out-experiment}, reveal that, even with three gold references, the upper bound on standard evaluation metrics is limited, not exceeding 0.66 out of 1.0. This reflects the intrinsic variability in natural language, even in relatively constrained domains such as wound care (which is less variable than general health or dermatology queries). 

Interestingly, GPT-4o responses are shown to consistently achieve higher scores than gold responses across all automatic metrics, and even approach or surpass human experts measured against each other. For instance, in the English subset, GPT-4o outperforms gold references on BLEU, ROUGE-L, and GPT-4o-as-judge scores. A similar pattern holds in Chinese, with especially large gains on GPT-4o-as-judge and ROUGE-L. 

These results suggest that automatic metrics can sometimes favor fluent and consistent outputs generated by LLMs over diverse but equally valid human-authored references. This highlights the dual risks of both $overestimating$ system quality when responses align well with surface features, and $underestimating$ it when valid variation is penalized. 

These findings underscore the limitations of reference-based NLG metrics, reinforce the need for cautious interpretation of automatic evaluation scores, and highlight the importance of developing more robust and context-aware evaluation frameworks, particularly in sensitive domains like medicine. 


\section{Conclusions}

In this work, we present the MORQA dataset, with 16,041 expert judgements, and conduct a comprehensive analysis of state-of-the-art evaluation metrics. Our findings show that LLM-as-a-judge methods generally outperform traditional metrics on English datasets. In Chinese, while LLM-based evaluators still perform strongly, traditional metrics such as BERTScore and BLEURT remain competitive on some subsets. 

Differences across languages and across specific aspects of evaluation suggest that a single evaluation score often reflects a nonlinear combination of multiple underlying quality criteria. For instance, a response that is more complete does not necessarily receive a higher overall score, highlighting the complexity of human judgment in NLG evaluation. The ablation study underscored key limitations of reference-based evaluation. Gold-standard responses exhibited limited measurable agreement with peer references, using current automatic metrics; while GPT-4o outputs frequently achieved higher metric scores. The observed limitations reinforce the need to develop evaluation methodologies that are comprehensive, reliable, and context-sensitive, particularly for healthcare.

Fine-tuning evaluation models remains an important direction for future work. To support this, we release the MORQA dataset with validation and test subsets. As LLM weights become increasingly open and accessible, we hope that this dataset and benchmarks will facilitate the development of more effective evaluation models and encourage further research on real-world medical question answering tasks. 

\section*{Limitations}

The work covers several medical open-response question answering datasets, however is limited in both sample size and scope. Only several human domain experts were employed to give ratings and their judgments may not fully represent all experts. Incorporating additional expert evaluations is an important area for expansion. Additionally, while we aimed to select representative models, not all LLMs or NLG evaluation methods were included in this analysis. 

Human evaluators included medical domain experts hired through Upwork as well as collaborating scientists and graduate trainees. Upwork annotators were selected based on verified professional credentials, provided with task descriptions and sample data, and asked to submit cost estimates. Compensation and workload were agreed upon prior to task initiation, ensuring fair pay and informed consent. 

The underlying QA datasets used are publicly available datasets licensed for research use. All newly created data, including model responses and domain expert ratings, were approved for public release and research use by the authors' institutions.

While automatic evaluation metrics provide a scalable means of estimating system performance, they are not a substitute for expert human judgment, particularly in sensitive domains such as healthcare. Human evaluation remains essential in sensitive settings. 

Finally, this study focuses only on English and Chinese datasets. The findings may not generalize to other languages and regions. Moreover, several models evaluated here depend on commercial LLM APIs or access to GPU compute resources, which may not be available to researchers in lower-resource settings. 




\bibliographystyle{plain}
\bibliography{main}

\begin{thebibliography}{10}

\bibitem{LiveMedQA2017}
Asma {Ben Abacha}, Eugene Agichtein, Yuval Pinter, and Dina Demner{-}Fushman.
\newblock Overview of the medical question answering task at trec 2017 liveqa.
\newblock In {\em TREC 2017}, 2017.

\bibitem{vqaonline}
Chongyan Chen, Mengchen Liu, Noel Codella, Yunsheng Li, Lu~Yuan, and Danna Gurari.
\newblock Fully authentic visual question answering dataset from online communities.
\newblock In Ales Leonardis, Elisa Ricci, Stefan Roth, Olga Russakovsky, Torsten Sattler, and G{\"{u}}l Varol, editors, {\em Computer Vision - {ECCV} 2024 - 18th European Conference, Milan, Italy, September 29-October 4, 2024, Proceedings, Part {XLVIII}}, volume 15106 of {\em Lecture Notes in Computer Science}, pages 252--269. Springer, 2024.

\bibitem{DeepSeekAI2024DeepSeekV3TR}
DeepSeek-AI, Aixin Liu, Bei Feng, Bing Xue, Bing-Li Wang, Bochao Wu, Chengda Lu, Chenggang Zhao, Chengqi Deng, Chenyu Zhang, Chong Ruan, Damai Dai, Daya Guo, Dejian Yang, Deli Chen, Dong-Li Ji, Erhang Li, Fangyun Lin, Fucong Dai, Fuli Luo, Guangbo Hao, Guanting Chen, Guowei Li, H.~Zhang, Han Bao, Hanwei Xu, Haocheng Wang, Haowei Zhang, Honghui Ding, Huajian Xin, Huazuo Gao, Hui Li, Hui Qu, J.~L. Cai, Jian Liang, Jianzhong Guo, Jiaqi Ni, Jiashi Li, Jiawei Wang, Jin Chen, Jingchang Chen, Jingyang Yuan, Junjie Qiu, Junlong Li, Jun-Mei Song, Kai Dong, Kai Hu, Kaige Gao, Kang Guan, Kexin Huang, Kuai Yu, Lean Wang, Lecong Zhang, Lei Xu, Leyi Xia, Liang Zhao, Litong Wang, Liyue Zhang, Meng Li, Miaojun Wang, Mingchuan Zhang, Minghua Zhang, Minghui Tang, Mingming Li, Ning Tian, Panpan Huang, Peiyi Wang, Peng Zhang, Qiancheng Wang, Qihao Zhu, Qinyu Chen, Qiushi Du, R.~J. Chen, R.~L. Jin, Ruiqi Ge, Ruisong Zhang, Ruizhe Pan, Runji Wang, Runxin Xu, Ruoyu Zhang, Ruyi Chen, S.~S. Li, Shanghao Lu, Shangyan Zhou, Shanhuang
  Chen, Shao-Ping Wu, Shengfeng Ye, Shirong Ma, Shiyu Wang, Shuang Zhou, Shuiping Yu, Shunfeng Zhou, Shuting Pan, T.~Wang, Tao Yun, Tian Pei, Tianyu Sun, W.~L. Xiao, Wangding Zeng, Wanjia Zhao, Wei An, Wen Liu, Wenfeng Liang, Wenjun Gao, Wen-Xuan Yu, Wentao Zhang, X.~Q. Li, Xiangyu Jin, Xianzu Wang, Xiaoling Bi, Xiaodong Liu, Xiaohan Wang, Xi-Cheng Shen, Xiaokang Chen, Xiaokang Zhang, Xiaosha Chen, Xiaotao Nie, Xiaowen Sun, Xiaoxiang Wang, Xin Cheng, Xin Liu, Xin Xie, Xingchao Liu, Xingkai Yu, Xinnan Song, Xinxia Shan, Xinyi Zhou, Xinyu Yang, Xinyuan Li, Xuecheng Su, Xuheng Lin, Y.~K. Li, Y.~Q. Wang, Y.~X. Wei, Y.~X. Zhu, Yang Zhang, Yanhong Xu, Yanping Huang, Yao Li, Yao Zhao, Yaofeng Sun, Yao Li, Yaohui Wang, Yi~Yu, Yi~Zheng, Yichao Zhang, Yifan Shi, Yi~Xiong, Ying He, Ying Tang, Yishi Piao, Yisong Wang, Yixuan Tan, Yi-Bing Ma, Yiyuan Liu, Yongqiang Guo, Yu~Wu, Yuan Ou, Yuchen Zhu, Yuduan Wang, Yue Gong, Yuheng Zou, Yujia He, Yukun Zha, Yunfan Xiong, Yunxiang Ma, Yuting Yan, Yu-Wei Luo, Yu~mei You, Yuxuan
  Liu, Yuyang Zhou, Z.~F. Wu, Zehui Ren, Zehui Ren, Zhangli Sha, Zhe Fu, Zhean Xu, Zhen Huang, Zhen Zhang, Zhenda Xie, Zhen guo Zhang, Zhewen Hao, Zhibin Gou, Zhicheng Ma, Zhigang Yan, Zhihong Shao, Zhipeng Xu, Zhiyu Wu, Zhongyu Zhang, Zhuoshu Li, Zihui Gu, Zijia Zhu, Zijun Liu, Zi-An Li, Ziwei Xie, Ziyang Song, Ziyi Gao, and Zizheng Pan.
\newblock Deepseek-v3 technical report.
\newblock {\em ArXiv}, abs/2412.19437, 2024.

\bibitem{Fu2023GPTScoreEA}
Jinlan Fu, See-Kiong Ng, Zhengbao Jiang, and Pengfei Liu.
\newblock Gptscore: Evaluate as you desire.
\newblock In {\em North American Chapter of the Association for Computational Linguistics}, 2023.

\bibitem{galley-etal-2015-deltableu}
Michel Galley, Chris Brockett, Alessandro Sordoni, Yangfeng Ji, Michael Auli, Chris Quirk, Margaret Mitchell, Jianfeng Gao, and Bill Dolan.
\newblock delta{BLEU}: A discriminative metric for generation tasks with intrinsically diverse targets.
\newblock In {\em Proceedings of the 53rd Annual Meeting of the Association for Computational Linguistics and the 7th International Joint Conference on Natural Language Processing (Volume 2: Short Papers)}, pages 445--450, Beijing, China, July 2015. Association for Computational Linguistics.

\bibitem{gupta2025carmodynamiccriteriageneration}
Taneesh Gupta, Shivam Shandilya, Xuchao Zhang, Rahul Madhavan, Supriyo Ghosh, Chetan Bansal, Huaxiu Yao, and Saravan Rajmohan.
\newblock Carmo: Dynamic criteria generation for context-aware reward modelling, 2025.

\bibitem{Kim2023PrometheusIF}
Seungone Kim, Jamin Shin, Yejin Cho, Joel Jang, S.~Longpre, Hwaran Lee, Sangdoo Yun, Seongjin Shin, Sungdong Kim, James Thorne, and Minjoon Seo.
\newblock Prometheus: Inducing fine-grained evaluation capability in language models.
\newblock {\em ArXiv}, abs/2310.08491, 2023.

\bibitem{Kim2024Prometheus2A}
Seungone Kim, Juyoung Suk, Shayne Longpre, Bill~Yuchen Lin, Jamin Shin, Sean Welleck, Graham Neubig, Moontae Lee, Kyungjae Lee, and Minjoon Seo.
\newblock Prometheus 2: An open source language model specialized in evaluating other language models.
\newblock {\em ArXiv}, abs/2405.01535, 2024.

\bibitem{lau2018dataset}
Jason~J Lau, Soumya Gayen, Asma Ben~Abacha, and Dina Demner-Fushman.
\newblock A dataset of clinically generated visual questions and answers about radiology images.
\newblock {\em Scientific data}, 5(1):1--10, 2018.

\bibitem{Lin2004ROUGEAP}
Chin-Yew Lin.
\newblock Rouge: A package for automatic evaluation of summaries.
\newblock In {\em Annual Meeting of the Association for Computational Linguistics}, 2004.

\bibitem{lin-2004-rouge}
Chin-Yew Lin.
\newblock {ROUGE}: A package for automatic evaluation of summaries.
\newblock In {\em Text Summarization Branches Out}, pages 74--81, Barcelona, Spain, July 2004. Association for Computational Linguistics.

\bibitem{Liu2020MedDGAE}
Wenge Liu, Jianheng Tang, Yi~Cheng, Wenjie Li, Yefeng Zheng, and Xiaodan Liang.
\newblock Meddg: An entity-centric medical consultation dataset for entity-aware medical dialogue generation.
\newblock In {\em Natural Language Processing and Chinese Computing}, 2020.

\bibitem{Papineni2002BleuAM}
Kishore Papineni, Salim Roukos, Todd Ward, and Wei-Jing Zhu.
\newblock Bleu: a method for automatic evaluation of machine translation.
\newblock In {\em Annual Meeting of the Association for Computational Linguistics}, 2002.

\bibitem{Sellam2020BLEURTLR}
Thibault Sellam, Dipanjan Das, and Ankur~P. Parikh.
\newblock Bleurt: Learning robust metrics for text generation.
\newblock In {\em Annual Meeting of the Association for Computational Linguistics}, 2020.

\bibitem{Srba2016ACS}
Ivan Srba and M{\'a}ria Bielikov{\'a}.
\newblock A comprehensive survey and classification of approaches for community question answering.
\newblock {\em ACM Transactions on the Web (TWEB)}, 10:1 -- 63, 2016.

\bibitem{Tian2017HowTM}
Zhiliang Tian, Rui Yan, Lili Mou, Yiping Song, Yansong Feng, and Dongyan Zhao.
\newblock How to make context more useful? an empirical study on context-aware neural conversational models.
\newblock In {\em Annual Meeting of the Association for Computational Linguistics}, 2017.

\bibitem{YIM2025104888}
Wen wai Yim, Asma {Ben Abacha}, Robert Doerning, Chia-Yu Chen, Jiaying Xu, Anita Subbarao, Zixuan Yu, Fei Xia, M.~Kennedy Hall, and Meliha Yetisgen.
\newblock Woundcarevqa: A multilingual visual question answering benchmark dataset for wound care.
\newblock {\em Journal of Biomedical Informatics}, 170:104888, 2025.

\bibitem{Ye2023FLASKFL}
Seonghyeon Ye, Doyoung Kim, Sungdong Kim, Hyeonbin Hwang, Seungone Kim, Yongrae Jo, James Thorne, Juho Kim, and Minjoon Seo.
\newblock Flask: Fine-grained language model evaluation based on alignment skill sets.
\newblock {\em ArXiv}, abs/2307.10928, 2023.

\bibitem{Yim2024DermaVQAAM}
Wen{-}wai Yim, Yujuan Fu, Zhaoyi Sun, Asma {Ben Abacha}, Meliha Yetisgen, and Fei Xia.
\newblock Dermavqa: {A} multilingual visual question answering dataset for dermatology.
\newblock In Marius~George Linguraru, Qi~Dou, Aasa Feragen, Stamatia Giannarou, Ben Glocker, Karim Lekadir, and Julia~A. Schnabel, editors, {\em Medical Image Computing and Computer Assisted Intervention - {MICCAI} 2024 - 27th International Conference, Marrakesh, Morocco, October 6-10, 2024, Proceedings, Part {V}}, volume 15005 of {\em Lecture Notes in Computer Science}, pages 209--219. Springer, 2024.

\bibitem{Yuan2021BARTScoreEG}
Weizhe Yuan, Graham Neubig, and Pengfei Liu.
\newblock Bartscore: Evaluating generated text as text generation.
\newblock {\em ArXiv}, abs/2106.11520, 2021.

\bibitem{zeng-etal-2020-meddialog}
Guangtao Zeng, Wenmian Yang, Zeqian Ju, Yue Yang, Sicheng Wang, Ruisi Zhang, Meng Zhou, Jiaqi Zeng, Xiangyu Dong, Ruoyu Zhang, Hongchao Fang, Penghui Zhu, Shu Chen, and Pengtao Xie.
\newblock {M}ed{D}ialog: Large-scale medical dialogue datasets.
\newblock In {\em Proceedings of the 2020 Conference on Empirical Methods in Natural Language Processing (EMNLP)}, pages 9241--9250, Online, November 2020. Association for Computational Linguistics.

\bibitem{Zhang2019ReCoSaDT}
Hainan Zhang, Yanyan Lan, Liang Pang, J.~Guo, and Xueqi Cheng.
\newblock Recosa: Detecting the relevant contexts with self-attention for multi-turn dialogue generation.
\newblock In {\em Annual Meeting of the Association for Computational Linguistics}, 2019.

\bibitem{Zhang2019BERTScoreET}
Tianyi Zhang, Varsha Kishore, Felix Wu, Kilian~Q. Weinberger, and Yoav Artzi.
\newblock Bertscore: Evaluating text generation with bert.
\newblock {\em ArXiv}, abs/1904.09675, 2019.

\bibitem{Zhao2019MoverScoreTG}
Wei Zhao, Maxime Peyrard, Fei Liu, Yang Gao, Christian~M. Meyer, and Steffen Eger.
\newblock Moverscore: Text generation evaluating with contextualized embeddings and earth mover distance.
\newblock In {\em Conference on Empirical Methods in Natural Language Processing}, 2019.

\bibitem{Zheng2023JudgingLW}
Lianmin Zheng, Wei-Lin Chiang, Ying Sheng, Siyuan Zhuang, Zhanghao Wu, Yonghao Zhuang, Zi~Lin, Zhuohan Li, Dacheng Li, Eric~P. Xing, Haotong Zhang, Joseph~E. Gonzalez, and Ion Stoica.
\newblock Judging llm-as-a-judge with mt-bench and chatbot arena.
\newblock {\em ArXiv}, abs/2306.05685, 2023.

\end{thebibliography}


\appendix

\section{Appendix}

\label{sec:appendix}

\begin{table}
  \centering
  \begin{tabular}{c}
    \hline
    \textbf{configurations} \\ \hline
    \underline{BERTSCORE} \\
    github.com/Tiiiger/bert\_score \\
    tokenizer: ''en''/''zh'' \\
    \underline{BLEURT} \\
    https://github.com/google-research/bleurt\\
    model:BLEURT-20\\
    \underline{BLEU} \\
     github.com/mjpost/sacrebleu\\
    use\_effective\_order: true\\
    EN tokenizer: tokenize\_13a\\
     ZH tokenizer: tokenize\_zh\\
    \underline{DEEPSEEK} \\
    ai.azure.com\\
    AZURE AI Foundry model ID:\\DeepSeek-V3-0324\\
    content-filter: None\\
    \underline{Gemini} \\
    https://cloud.google.com/ai/generative-ai?hl=en\\
    Google GenAI Model Name:\\gemini-1.5-pro-002\\
    hate-speech: Block-None\\
    harrassment: Block-None\\
    \underline{GPT-4o} \\
    oai.azure.com\\
    AZURE AI Foundry OpenAI model name:\\gpt-4o\\
    content-filter: None\\
    \underline{PROMETHEUS}\\
    ai.azure.com\\
    AZURE AI Foundry model ID:\\prometheus-eval-prometheus-7b-v2.0\\
    \underline{ROUGE}\\
    huggingface.co/spaces/evaluate-metric/rouge \\
    tokenizer: same as BLEU\\
    \hline
  \end{tabular}
  \caption{Evaluation Metric Configurations. Defaults are used if not otherwise mentioned.}
  \label{tab:eval-settings}
\end{table}

\begin{table*}
  \centering
  \scalebox{.8}{
  \begin{tabular}{p{14cm}}
    \hline
    [INST] You are a fair judge assistant tasked with providing clear, objective feedback based on specific criteria, ensuring each assessment reflects the absolute standards set for performance.\\
    \#\#\#Task Description:\\
    An instruction (might include an Input inside it), a response to evaluate, several reference answers that get a score of 1, and a score rubric representing a evaluation criteria are given.\\
    1. Write a detailed feedback that assess the quality of the response strictly based on the given score rubric, not evaluating in general.\\
    2. After writing a feedback, write a score that is either 0, 0.5, or 1. You should refer to the score rubric.\\
    3. The output format should look as follows: "Feedback: (write a feedback for criteria) $[$RESULT$]$ (one of the options 0, 0.5, 1)"\\
    4. Please do not generate any other opening, closing, and explanations.\\
    \#\#\#The instruction to evaluate:\\
    Facing a wound-related medical problem, a patient seeks medical advice from a doctor. Their QUERY to the doctor is the following. \{\} \\
    \#\#\#Response to evaluate:\{\}\\
    \#\#\#Reference Answers (Score 1):\{\}\\
    \#\#\#Score Rubrics:\\
    $[$Is the model proficient in applying medical factual accuracy, appropriate medical advice, and patient-understandable language to its responses?$]$\\
    Score 0: The model neglects to provide a complete appropriate response, or the response includes medically inaccurate advice.\\
    Score 0.5: The model answers with some relevant medically accurate advice, however the advice is incomplete.\\
    Score 1: The model answers with a medically accurate and appropriate, complete response.\\
    \#\#\#Feedback:  $[$/INST$]$\\
    \hline
    (The Chinese prompt is same as English prompt, except for the following sections)\\
    \#\#\#The instruction to evaluate:\\
    \chinese{面对一个有关伤口的医疗问题，病人寻求医生提供医疗建议。以下是他给医生的提问: 伤口拆线后 伤口附近怎么起皮啊: }\\
    \#\#\#Score Rubrics:\\
    $[$Is the model proficient in applying medical factual consistency and patient-understandable language to its responses, given a list of Reference Answers?$]$\\
Score 0: The model neglects to provide a complete appropriate response, or the response is factually inconsistent with all Reference Answers.\\
Score 0.5: The model answers with incomplete advise, but is factually consistent with at least one of the Reference Answers.\\
Score 1: The model answers with complete advise and is factually consistent with at least one of the Reference Answers.\\
    \hline
  \end{tabular}}
  \caption{Prometheus Prompt (English TOP, Chinese BOTTOM)}
  \label{tab:prometheus-prompts}
\end{table*}

\begin{table*}
  \centering
  \scalebox{.8}{
  \begin{tabular}{p{14cm}}
    \hline
    SYSTEM: You are a helpful medical assistant.\\
    USER: Given a patient \{QUERY\}, and a list of \{REFERENCE RESPONSES\}, please evaluate a \{CANDIDATE RESPONSE\} using a three-step rating described below.\\
Rating: 0 - \{CANDIDATE RESPONSE\} is incomplete and may contain medically incorrect advice.\\
Rating: 0.5 - \{CANDIDATE RESPONSE\} is incomplete but has partially correct medical advice.\\
Rating: 1.0 - \{CANDIDATE RESPONSE\} is complete and has medically correct advice.\\
The \{REFERENCE RESPONSES\} represent answers given by domain experts and can be used as references for evaluation.\\
QUERY:\\
REFERENCE RESPONSES:\\
CANDIDATE RESPONSE:\\
RATING:\\
    \hline
\chinese{SYSTEM: 你是一个很有帮助的医疗助手。}\\
\chinese{USER: 给定病人提出的\{问题\}以及一系列\{参考回复\}，请使用下述的３级评分制度来评估\{待测回复\}。}\\
\chinese{评分： 0 - \{待测回复\} 不完整且与所有\{参考回复\}事实不符。}\\
\chinese{评分：0.5 - \{待测回复\} 不完整但与至少一个\{参考回复\}事实相符。}\\
\chinese{评分：1.0 - \{待测回复\} 完整且与至少一个\{参考回复\}事实相符}\\
\chinese{问题: \{\}}\\
\chinese{参考回复:\{\}}\\
\chinese{待测回复:\{\}}\\
\chinese{评分:}\\
    \hline
  \end{tabular}
  }
  \caption{LLM-as-Judge Prompts for General AI Models (English TOP, Chinese BOTTOM)}
  \label{tab:prompts}
\end{table*}

\begin{table*}
  \centering
  \scalebox{.8}{
  \begin{tabular}{p{15cm}}
    \hline
    EN \\ \hline
    To evaluate the **CANDIDATE RESPONSE**, let's break it down step-by-step against the **REFERENCE RESPONSES** and the **QUERY**:\\\\\#\#\# **Key Points from the Query:**\\- A previously healed wound on the sole reopened due to scratching and friction.  \\\#\#\# **Key Points from Reference Responses:**\\1. **Possible diagnoses:** Corn/callus or plantar wart (suggesting OTC treatments like salicylic acid, soaking, pumice stone, or duct tape).  \\2. **General advice:**  \\   - Cover the area to reduce friction.  \\   - Replace worn-out shoes.  \\   - Seek podiatry evaluation if no improvement.  \\3. **Caution:** Hard to confirm diagnosis without in-person evaluation (possible foreign body or infection risk).  \\\\\#\#\# **Candidate Response Evaluation:**\\- **Strengths:**  \\  - Advises keeping the wound **clean and covered** (matches reference advice).  \\  - Recommends **reducing friction** (cushioned insoles align with replacing shoes/insoles).  \\  - Suggests **OTC pain relief** (reasonable for symptom management).  \\  - Advises **medical consultation if worsening** (consistent with references).  \\- **Weaknesses:**  \\  - Does **not** suggest specific OTC treatments (e.g., salicylic acid, soaking, pumice stone) mentioned in references.  \\  - Does **not** propose possible diagnoses (e.g., corn, wart) or tailored treatments.  \\  - Lacks emphasis on **shoe replacement** (a key friction-reduction step in references).  \\\\\#\#\# **Rating: 0.5**  \\- The response is **partially correct and incomplete**.  \\- It covers **basic wound care and pain management** but misses **specific treatments and diagnostic considerations** from references.  \\- It does not contain **medically incorrect advice**, but it lacks completeness.  \\\\**Reasoning:**  \\- **0.5** = Incomplete but partially correct (general advice is safe, but lacks specificity).  \\- **Not 1.0** because it omits key details and possible diagnoses.  \\- **Not 0** because it provides some valid, non-harmful guidance.  \\\\**Suggestions for Improvement:**  \\- Add possible causes (e.g., "This may be a corn or plantar wart; consider OTC salicylic acid or soaking.). \\
    \hline
    ZH \\ \hline
    \chinese{评分: 1.0}\\ \\ \chinese{解释: {待测回复} 完整且与至少一个{参考回复}事实相符。具体来说，待测回复与最后一个参考回复几乎完全一致，提供了完整的诊断信息（线状苔藓、伊藤色素减少症或色素失禁症的色素减退性皮损）和适当的处理建议（观察，无需积极干预）。因此，评分为1.0。} \\
    \hline
  \end{tabular}
  }
  \caption{DeepSeekV3 LLM-as-judge Rationale Example}
  \label{tab:rationaleacc}
\end{table*}

\begin{table*}
  \centering
  \scalebox{.8}{
  \begin{tabular}{p{15cm}}
    \hline
    EN \\ \hline
    The response provided is accurate and relevant to the patient's query. It correctly identifies the condition as tinea corporis, which is a type of fungal infection commonly known as ringworm. The advice to see a doctor for a proper diagnosis and treatment is appropriate and patient-understandable. Additionally, the response mentions that the treatment is usually easy, which is a helpful piece of information for the patient. However, the response could have been more complete by providing more details about the treatment, such as the type of antifungal medication that might be used. Despite this, the response still meets the criteria for a score of 1 as it is medically accurate, appropriate, and complete. So the overall score is 1. [RESULT] 1 \\ \hrule
    ZH \\ \hrule
    The response provided is factually consistent with the reference answer, which correctly identifies the condition as line-shaped eczema. The advice given is also complete, covering the use of gentle cleansers, the avoidance of harsh treatments, and the need for further examination. However, the response does not mention the possibility of color reduction or the specific location of the skin damage, which are important details in the reference answer. This omission results in a less comprehensive response compared to the reference answer. Therefore, while the response is accurate and patient-friendly, it lacks the depth and detail found in the reference answer. So the overall score is 1. [RESULT] 1\\\hrule
  \end{tabular}
  }
  \caption{Prometheus LLM-as-judge Rationale Example}
  \label{tab:rationaleacc2}
\end{table*}

\begin{table*}
  \centering
  \scalebox{.8}{
  \begin{tabular}{p{15cm}}
    \hline
    EN \\ \hline
    RATING: 0\\\\The candidate response is incomplete and provides insufficient medical advice. It suggests that the condition could be parapsoriasis and recommends seeing a dermatologist, but does not offer further guidance or details that align with the reference responses, which primarily indicate a high likelihood of psoriasis and suggest specific treatments and examinations. Furthermore, it does not recommend a pathological examination which is critical for confirming the diagnosis.\\
    \\ \hline
    ZH \\ \hline
    \chinese{评分：1.0}\\\\\chinese{评估理由：}\\\chinese{待测回复为“这是线状苔藓，一种良性疾病。它无害，也不会传染。使用温和、无香味的肥皂和清洁剂。避免用力擦洗或任何可能加剧病情的刺激物。需要复诊”。}\\\chinese{该回复与参考回复中指出的线状苔藓作为可能的诊断是一致的，且完整涵盖了对这一情况的解释和建议。因此，评分为1.0。}
    \hrule
  \end{tabular}
  }
  \caption{GPT-4O LLM-as-judge Rationale Example}
  \label{tab:rationaleacc3}
\end{table*}

\begin{table*}
  \centering
  \scalebox{.8}{
  \begin{tabular}{p{15cm}}
    \hline
    EN \\ \hline
    RATING: 0.5\\\\The candidate response acknowledges the possibility of Parapsoriasis, but several reference responses suggest Psoriasis or other conditions as more likely.  While recommending a dermatologist visit is crucial and correct, the response doesn't offer any further information or direction.  It sits on the fence without providing more specific guidance based on the symptoms described.  Therefore, it's incomplete but contains the partially correct advice of seeking expert consultation.\\
    \\ \hline
    ZH \\ \hline
    \chinese{评分：1.0}\\\\ \chinese{待测回复完整地描述了线状苔藓的良性特质、不传染性，并提供了护理建议和复诊建议。这与多个参考回复中的信息相符，例如“考虑是线状苔藓，观察就可以”，“这是线状苔藓，一种良性疾病。它无害，也不会传染……”等。}
    \hrule
  \end{tabular}
  }
  \caption{Gemini LLM-as-judge Rationale Example}
  \label{tab:rationaleacc4}
\end{table*}

\begin{table*} 
\centering 
\scalebox{.65}{ 
\begin{tabular}{c|ccc|ccc|ccc} 
\hline 
 & & DermaVQA & & & LiveQA & & & WoundcareVQA & \\
\hline 
metric  & $\tau$ & $r$ & $\rho$ & $\tau$ & r & $\rho$ & $\tau$ & $r$ & $\rho$ \\ \hline
bertscore-f1-max & 0.00 & 0.00 & 0.00 & 0.00 & 0.00 & 0.00 & 0.00 & 0.00 & 0.00 \\
bertscore-f1-mean & 0.00 & 0.00 & 0.00 & 0.00 & 0.00 & 0.00 & 0.00 & 0.00 & 0.00 \\
bleurt20-max & 0.00 & 0.00 & 0.00 & 0.23 & 0.26 & 0.23 & 0.00 & 0.00 & 0.00 \\
bleurt20-mean & 0.00 & 0.00 & 0.00 & 0.18 & 0.26 & 0.19 & 0.00 & 0.00 & 0.00 \\
deepseekv3 & 0.00 & 0.00 & 0.00 & 0.00 & 0.00 & 0.00 & 0.00 & 0.00 & 0.00 \\
deltableu-micro & 0.00 & 0.00 & 0.00 & 0.00 & 0.00 & 0.00 & 0.00 & 0.00 & 0.00 \\
gpt-4o & 0.00 & 0.00 & 0.00 & 0.00 & 0.00 & 0.00 & 0.00 & 0.00 & 0.00 \\
gemini-1.5-pro & 0.00 & 0.00 & 0.00 & 0.00 & 0.00 & 0.00 & 0.00 & 0.00 & 0.00 \\
promethesus7bv2 & 0.00 & 0.00 & 0.00 & 0.00 & 0.00 & 0.00 & 0.00 & 0.00 & 0.00 \\
rouge-1 & 0.25 & 0.01 & 0.26 & 0.03 & 0.12 & 0.03 & 0.23 & 0.00 & 0.22 \\
rouge-2 & 0.00 & 0.00 & 0.00 & 0.00 & 0.00 & 0.00 & 0.00 & 0.00 & 0.00 \\
rouge-L & 0.00 & 0.00 & 0.00 & 0.01 & 0.08 & 0.01 & 0.00 & 0.00 & 0.00 \\
\hline
\end{tabular}}
\caption{woundcare-en-overall p-val ($\tau$=kendall's tau, r=pearson, $\rho$=spearman) 0.00 means p-values of $<0.01$.  }
\label{tab:woundcare-en-overall-pval}
\end{table*}

\begin{table*} 
\centering 
\scalebox{.65}{ 
\begin{tabular}{c|ccc|ccc|ccc} 
\hline 
 & & DermaVQA & & & Meddialog  & & & WoundcareVQA & \\
\hline 
metric  & $\tau$ & $r$ & $\rho$ & $\tau$ & r & $\rho$ & $\tau$ & $r$ & $\rho$ \\ \hline
bertscore-f1-max & 0.00 & 0.00 & 0.00 & 0.00 & 0.00 & 0.00 & 0.04 & 0.03 & 0.04 \\
bertscore-f1-mean & 0.00 & 0.00 & 0.00 & 0.00 & 0.00 & 0.00 & 0.04 & 0.02 & 0.04 \\
bleurt20-max & 0.00 & 0.00 & 0.00 & 0.00 & 0.00 & 0.00 & 0.00 & 0.00 & 0.00 \\
bleurt20-mean & 0.00 & 0.00 & 0.00 & 0.00 & 0.00 & 0.00 & 0.00 & 0.00 & 0.00 \\
deepseekv3 & 0.00 & 0.00 & 0.00 & 0.00 & 0.00 & 0.00 & 0.00 & 0.00 & 0.00 \\
deltableu-micro & 0.00 & 0.00 & 0.00 & 0.00 & 0.00 & 0.00 & 0.04 & 0.06 & 0.03 \\
gpt-4o & 0.00 & 0.00 & 0.00 & 0.00 & 0.00 & 0.00 & 0.00 & 0.00 & 0.00 \\
gemini-1.5-pro & 0.00 & 0.00 & 0.00 & 0.00 & 0.00 & 0.00 & 0.00 & 0.00 & 0.00 \\
promethesus7bv2 & 0.01 & 0.03 & 0.01 & 0.00 & 0.00 & 0.00 & 0.24 & 0.10 & 0.24 \\
rouge-1 & 0.00 & 0.00 & 0.00 & 0.00 & 0.00 & 0.00 & 0.33 & 0.26 & 0.33 \\
rouge-2 & 0.00 & 0.00 & 0.00 & 0.00 & 0.00 & 0.00 & 0.01 & 0.01 & 0.01 \\
rouge-L & 0.04 & 0.01 & 0.04 & 0.00 & 0.00 & 0.00 & 0.80 & 0.77 & 0.82 \\
\hline
\end{tabular}}
\caption{woundcare-zh-factual-consistency-wgold p-val ($\tau$=kendall's tau, r=pearson, $\rho$=spearman). 0.00 means p-values of $<0.01$. }
\label{tab:woundcare-zh-factual-consistency-wgold-pval}
\end{table*}

\begin{table*} 
\centering 
\begin{tabular}{cccccc} 
\hline 
metric  & kendall & pearson & spearman & avg-corr & acc \\ \hline
deepseekv3 & -0.3054 & -0.3400 & -0.3216 & -0.3223 & 0.4753 \\ 
gpt-4o & -0.3052 & -0.3238 & -0.3122 & -0.3138 & 0.4910 \\ 
gemini-1.5-pro & -0.3010 & -0.3131 & -0.3049 & -0.3063 & 0.4875 \\ 
bleurt20-max & -0.1702 & -0.2146 & -0.2081 & -0.1977 & 0.3290 \\ 
bertscore-f1-max & -0.1364 & -0.1675 & -0.1667 & -0.1569 & 0.2975 \\ 
deltableu-micro & -0.1483 & -0.1367 & -0.1813 & -0.1555 & 0.3032 \\ 
bleurt20-mean & -0.1284 & -0.1680 & -0.1570 & -0.1511 & 0.3147 \\ 
promethesus7bv2 & -0.1226 & -0.1472 & -0.1256 & -0.1318 & 0.5297 \\ 
bertscore-f1-mean & -0.1115 & -0.1342 & -0.1363 & -0.1273 & 0.2860 \\ 
rouge-L & -0.0897 & -0.1530 & -0.1096 & -0.1175 & 0.2889 \\ 
rouge-2 & -0.0786 & -0.1305 & -0.0961 & -0.1018 & 0.3032 \\ 
rouge-1 & -0.0221 & -0.0956 & -0.0270 & -0.0482 & 0.3520 \\ 
\hline
\end{tabular}
\caption{woundcare-en-disagree\_flag}
\label{tab:woundcare-en-disagree\_flag}
\end{table*}

\begin{table*} 
\centering 
\begin{tabular}{cccccc} 
\hline 
metric  & kendall & pearson & spearman & avg-corr & acc \\ \hline
deepseekv3 & 0.2905 & 0.3377 & 0.3068 & 0.3117 & 0.6100 \\ 
gemini-1.5-pro & 0.2720 & 0.3162 & 0.2758 & 0.2880 & 0.6409 \\ 
gpt-4o & 0.2733 & 0.3079 & 0.2804 & 0.2872 & 0.6108 \\ 
promethesus7bv2 & 0.1913 & 0.2433 & 0.1975 & 0.2107 & 0.5986 \\ 
bertscore-f1-mean & 0.1707 & 0.2501 & 0.2092 & 0.2100 & 0.3692 \\ 
bleurt20-max & 0.1672 & 0.2571 & 0.2051 & 0.2098 & 0.3348 \\ 
bertscore-f1-max & 0.1705 & 0.2378 & 0.2092 & 0.2058 & 0.3462 \\ 
bleurt20-mean & 0.1574 & 0.2640 & 0.1931 & 0.2048 & 0.3462 \\ 
rouge-L & 0.1373 & 0.2373 & 0.1685 & 0.1810 & 0.3606 \\ 
rouge-2 & 0.0970 & 0.2558 & 0.1190 & 0.1573 & 0.3405 \\ 
rouge-1 & 0.0588 & 0.2251 & 0.0726 & 0.1188 & 0.3491 \\ 
deltableu-micro & 0.1003 & 0.1098 & 0.1233 & 0.1112 & 0.3491 \\ 
\hline
\end{tabular}
\caption{woundcare-en-completeness}
\label{tab:woundcare-en-completeness}
\end{table*}

\begin{table*} 
\centering 
\begin{tabular}{cccccc} 
\hline 
metric  & kendall & pearson & spearman & avg-corr & acc \\ \hline
gpt-4o & 0.3288 & 0.3704 & 0.3416 & 0.3470 & 0.5885 \\ 
deepseekv3 & 0.3150 & 0.3820 & 0.3390 & 0.3454 & 0.5814 \\ 
gemini-1.5-pro & 0.3056 & 0.3360 & 0.3146 & 0.3187 & 0.6065 \\ 
bleurt20-max & 0.1405 & 0.2172 & 0.1749 & 0.1775 & 0.3749 \\ 
bertscore-f1-mean & 0.1156 & 0.1906 & 0.1434 & 0.1499 & 0.4208 \\ 
bertscore-f1-max & 0.1101 & 0.1872 & 0.1377 & 0.1450 & 0.4007 \\ 
bleurt20-mean & 0.1034 & 0.1947 & 0.1288 & 0.1423 & 0.3864 \\ 
promethesus7bv2 & 0.1118 & 0.1476 & 0.1173 & 0.1256 & 0.5699 \\ 
deltableu-micro & 0.0897 & 0.0989 & 0.1124 & 0.1003 & 0.3921 \\ 
rouge-L & 0.0262 & 0.1717 & 0.0326 & 0.0769 & 0.4036 \\ 
rouge-1 & 0.0210 & 0.1703 & 0.0271 & 0.0728 & 0.3434 \\ 
rouge-2 & 0.0259 & 0.1532 & 0.0327 & 0.0706 & 0.3663 \\ 
\hline
\end{tabular}
\caption{woundcare-en-factual-accuracy}
\label{tab:woundcare-en-factual-accuracy}
\end{table*}

\begin{table*} 
\centering 
\begin{tabular}{cccccc} 
\hline 
metric  & kendall & pearson & spearman & avg-corr & acc \\ \hline
gpt-4o & 0.2420 & 0.2871 & 0.2479 & 0.2590 & 0.6466 \\ 
gemini-1.5-pro & 0.2217 & 0.2649 & 0.2248 & 0.2371 & 0.6323 \\ 
promethesus7bv2 & 0.2164 & 0.2512 & 0.2223 & 0.2300 & 0.6043 \\ 
deepseekv3 & 0.1967 & 0.2365 & 0.2071 & 0.2134 & 0.5728 \\ 
rouge-2 & 0.1356 & 0.3344 & 0.1659 & 0.2119 & 0.2688 \\ 
rouge-1 & 0.1149 & 0.3470 & 0.1409 & 0.2009 & 0.2717 \\ 
bertscore-f1-mean & 0.1480 & 0.2718 & 0.1810 & 0.2003 & 0.2717 \\ 
bleurt20-mean & 0.1317 & 0.3003 & 0.1615 & 0.1979 & 0.2631 \\ 
rouge-L & 0.1164 & 0.3288 & 0.1426 & 0.1960 & 0.2717 \\ 
bleurt20-max & 0.1465 & 0.2620 & 0.1794 & 0.1960 & 0.2631 \\ 
bertscore-f1-max & 0.1339 & 0.2294 & 0.1642 & 0.1759 & 0.2717 \\ 
deltableu-micro & 0.1031 & 0.1150 & 0.1263 & 0.1148 & 0.2659 \\ 
\hline
\end{tabular}
\caption{woundcare-en-relevance}
\label{tab:woundcare-en-relevance}
\end{table*}

\begin{table*} 
\centering 
\begin{tabular}{cccccc} 
\hline 
metric  & kendall & pearson & spearman & avg-corr & acc \\ \hline
bleurt20-mean & 0.1817 & 0.3563 & 0.2234 & 0.2538 & 0.3147 \\ 
bertscore-f1-mean & 0.1889 & 0.3383 & 0.2323 & 0.2531 & 0.3262 \\ 
bleurt20-max & 0.1930 & 0.3199 & 0.2369 & 0.2499 & 0.3204 \\ 
gemini-1.5-pro & 0.2156 & 0.2544 & 0.2199 & 0.2300 & 0.6122 \\ 
gpt-4o & 0.2073 & 0.2496 & 0.2137 & 0.2235 & 0.6358 \\ 
rouge-2 & 0.1202 & 0.3394 & 0.1458 & 0.2018 & 0.3118 \\ 
bertscore-f1-max & 0.1485 & 0.2596 & 0.1822 & 0.1968 & 0.3176 \\ 
promethesus7bv2 & 0.1719 & 0.2226 & 0.1773 & 0.1906 & 0.5957 \\ 
deepseekv3 & 0.1676 & 0.2101 & 0.1776 & 0.1851 & 0.5670 \\ 
rouge-L & 0.0852 & 0.3181 & 0.1047 & 0.1693 & 0.3090 \\ 
rouge-1 & 0.0709 & 0.3440 & 0.0866 & 0.1672 & 0.3061 \\ 
deltableu-micro & 0.1247 & 0.1502 & 0.1534 & 0.1428 & 0.3176 \\ 
\hline
\end{tabular}
\caption{woundcare-en-writing-style}
\label{tab:woundcare-en-writing-style}
\end{table*}

\begin{table*} 
\centering 
\begin{tabular}{cccccc} 
\hline 
metric  & kendall & pearson & spearman & avg-corr & acc \\ \hline
deepseekv3 & 0.3696 & 0.4310 & 0.3988 & 0.3998 & 0.5957 \\ 
gpt-4o & 0.3649 & 0.4060 & 0.3796 & 0.3835 & 0.5943 \\ 
gemini-1.5-pro & 0.3603 & 0.3798 & 0.3721 & 0.3707 & 0.6351 \\ 
bleurt20-max & 0.2157 & 0.2962 & 0.2698 & 0.2605 & 0.4065 \\ 
bertscore-f1-max & 0.1782 & 0.2515 & 0.2227 & 0.2175 & 0.4495 \\ 
promethesus7bv2 & 0.1866 & 0.2278 & 0.1961 & 0.2035 & 0.5928 \\ 
bleurt20-mean & 0.1594 & 0.2513 & 0.1997 & 0.2035 & 0.4265 \\ 
bertscore-f1-mean & 0.1613 & 0.2371 & 0.2015 & 0.2000 & 0.4609 \\ 
deltableu-micro & 0.1605 & 0.1687 & 0.2007 & 0.1766 & 0.4237 \\ 
rouge-2 & 0.1029 & 0.2317 & 0.1292 & 0.1546 & 0.3978 \\ 
rouge-L & 0.0993 & 0.2193 & 0.1247 & 0.1478 & 0.4409 \\ 
rouge-1 & 0.0401 & 0.2073 & 0.0523 & 0.0999 & 0.3864 \\ 
\hline
\end{tabular}
\caption{woundcare-en-overall}
\label{tab:woundcare-en-overall}
\end{table*}

\begin{table*} 
\centering 
\begin{tabular}{cccccc} 
\hline 
metric  & kendall & pearson & spearman & avg-corr & acc \\ \hline
gemini-1.5-pro & -0.3766 & -0.3835 & -0.3849 & -0.3817 & 0.5800 \\ 
deepseekv3 & -0.3364 & -0.3531 & -0.3488 & -0.3461 & 0.5413 \\ 
gpt-4o & -0.3283 & -0.3244 & -0.3370 & -0.3299 & 0.4980 \\ 
bleurt20-mean & -0.2416 & -0.2945 & -0.2962 & -0.2774 & 0.3147 \\ 
bleurt20-max & -0.2182 & -0.2310 & -0.2674 & -0.2389 & 0.3067 \\ 
bertscore-f1-mean & -0.1948 & -0.2545 & -0.2384 & -0.2292 & 0.2960 \\ 
bertscore-f1-max & -0.1366 & -0.1802 & -0.1679 & -0.1616 & 0.3147 \\ 
deltableu-micro & -0.1341 & -0.0955 & -0.1641 & -0.1312 & 0.3307 \\ 
rouge-L & -0.0789 & -0.1203 & -0.0965 & -0.0986 & 0.3573 \\ 
rouge-2 & -0.0710 & -0.1096 & -0.0870 & -0.0892 & 0.3707 \\ 
rouge-1 & -0.0252 & -0.1011 & -0.0305 & -0.0523 & 0.3413 \\ 
promethesus7bv2 & 0.0400 & 0.0460 & 0.0425 & 0.0429 & 0.4667 \\ 
\hline
\end{tabular}
\caption{iiyi-en-disagree\_flag}
\label{tab:iiyi-en-disagree\_flag}
\end{table*}

\begin{table*} 
\centering 
\begin{tabular}{cccccc} 
\hline 
metric  & kendall & pearson & spearman & avg-corr & acc \\ \hline
bleurt20-mean & 0.2434 & 0.3090 & 0.3086 & 0.2870 & 0.5200 \\ 
bertscore-f1-max & 0.2496 & 0.2891 & 0.3202 & 0.2863 & 0.5387 \\ 
bertscore-f1-mean & 0.2331 & 0.2833 & 0.2957 & 0.2707 & 0.5360 \\ 
bleurt20-max & 0.2345 & 0.2663 & 0.2986 & 0.2664 & 0.5253 \\ 
promethesus7bv2 & 0.2268 & 0.2306 & 0.2519 & 0.2365 & 0.5733 \\ 
gemini-1.5-pro & 0.2161 & 0.2492 & 0.2292 & 0.2315 & 0.6040 \\ 
gpt-4o & 0.2168 & 0.2420 & 0.2333 & 0.2307 & 0.5940 \\ 
deepseekv3 & 0.1591 & 0.1775 & 0.1718 & 0.1695 & 0.6107 \\ 
rouge-L & 0.1460 & 0.1680 & 0.1860 & 0.1667 & 0.4693 \\ 
rouge-2 & 0.1349 & 0.1752 & 0.1723 & 0.1608 & 0.4773 \\ 
deltableu-micro & 0.1221 & 0.0738 & 0.1564 & 0.1174 & 0.4640 \\ 
rouge-1 & 0.0515 & 0.0957 & 0.0652 & 0.0708 & 0.4587 \\ 
\hline
\end{tabular}
\caption{iiyi-en-completeness}
\label{tab:iiyi-en-completeness}
\end{table*}

\begin{table*} 
\centering 
\begin{tabular}{cccccc} 
\hline 
metric  & kendall & pearson & spearman & avg-corr & acc \\ \hline
gemini-1.5-pro & 0.3875 & 0.3812 & 0.4127 & 0.3938 & 0.6067 \\ 
bleurt20-mean & 0.3236 & 0.3618 & 0.4108 & 0.3654 & 0.5547 \\ 
bleurt20-max & 0.3215 & 0.3209 & 0.4072 & 0.3499 & 0.5600 \\ 
deepseekv3 & 0.3201 & 0.3205 & 0.3488 & 0.3298 & 0.5973 \\ 
bertscore-f1-mean & 0.2875 & 0.3112 & 0.3674 & 0.3221 & 0.5653 \\ 
gpt-4o & 0.3199 & 0.2920 & 0.3450 & 0.3190 & 0.5660 \\ 
bertscore-f1-max & 0.2774 & 0.2908 & 0.3554 & 0.3079 & 0.5600 \\ 
deltableu-micro & 0.2352 & 0.1620 & 0.3039 & 0.2337 & 0.5333 \\ 
rouge-L & 0.1827 & 0.1911 & 0.2328 & 0.2022 & 0.5040 \\ 
rouge-2 & 0.1188 & 0.1444 & 0.1528 & 0.1387 & 0.4693 \\ 
promethesus7bv2 & 0.1026 & 0.0851 & 0.1135 & 0.1004 & 0.5173 \\ 
rouge-1 & 0.0464 & 0.0746 & 0.0586 & 0.0599 & 0.4667 \\ 
\hline
\end{tabular}
\caption{iiyi-en-factual-accuracy}
\label{tab:iiyi-en-factual-accuracy}
\end{table*}

\begin{table*} 
\centering 
\begin{tabular}{cccccc} 
\hline 
metric  & kendall & pearson & spearman & avg-corr & acc \\ \hline
gemini-1.5-pro & 0.3846 & 0.3851 & 0.4084 & 0.3927 & 0.6147 \\ 
deepseekv3 & 0.3369 & 0.3414 & 0.3633 & 0.3472 & 0.5867 \\ 
gpt-4o & 0.3364 & 0.3108 & 0.3612 & 0.3361 & 0.6113 \\ 
bleurt20-mean & 0.2942 & 0.3316 & 0.3728 & 0.3329 & 0.5067 \\ 
bleurt20-max & 0.3017 & 0.2925 & 0.3816 & 0.3253 & 0.5307 \\ 
bertscore-f1-mean & 0.2821 & 0.3087 & 0.3601 & 0.3170 & 0.5387 \\ 
bertscore-f1-max & 0.2528 & 0.2604 & 0.3260 & 0.2798 & 0.5360 \\ 
deltableu-micro & 0.2123 & 0.1388 & 0.2731 & 0.2081 & 0.5013 \\ 
rouge-L & 0.1593 & 0.1702 & 0.2053 & 0.1783 & 0.4587 \\ 
rouge-2 & 0.1149 & 0.1383 & 0.1474 & 0.1335 & 0.4427 \\ 
rouge-1 & 0.0585 & 0.1015 & 0.0731 & 0.0777 & 0.4480 \\ 
promethesus7bv2 & 0.0679 & 0.0364 & 0.0741 & 0.0595 & 0.5013 \\ 
\hline
\end{tabular}
\caption{iiyi-en-relevance}
\label{tab:iiyi-en-relevance}
\end{table*}

\begin{table*} 
\centering 
\begin{tabular}{cccccc} 
\hline 
metric  & kendall & pearson & spearman & avg-corr & acc \\ \hline
bleurt20-mean & 0.2452 & 0.3094 & 0.3116 & 0.2887 & 0.4907 \\ 
gemini-1.5-pro & 0.2686 & 0.3065 & 0.2844 & 0.2865 & 0.6493 \\ 
gpt-4o & 0.2640 & 0.2982 & 0.2820 & 0.2814 & 0.5980 \\ 
deepseekv3 & 0.2279 & 0.2666 & 0.2460 & 0.2468 & 0.6400 \\ 
bleurt20-max & 0.2076 & 0.2158 & 0.2668 & 0.2300 & 0.4960 \\ 
bertscore-f1-mean & 0.1542 & 0.2290 & 0.1954 & 0.1929 & 0.4827 \\ 
bertscore-f1-max & 0.1548 & 0.2045 & 0.1983 & 0.1859 & 0.4880 \\ 
rouge-2 & 0.1397 & 0.1712 & 0.1777 & 0.1628 & 0.4587 \\ 
promethesus7bv2 & 0.1385 & 0.1234 & 0.1522 & 0.1380 & 0.5600 \\ 
rouge-L & 0.1067 & 0.1220 & 0.1361 & 0.1216 & 0.4400 \\ 
rouge-1 & 0.0647 & 0.1490 & 0.0822 & 0.0986 & 0.4480 \\ 
deltableu-micro & 0.0306 & -0.0325 & 0.0398 & 0.0127 & 0.4347 \\ 
\hline
\end{tabular}
\caption{iiyi-en-writing-style}
\label{tab:iiyi-en-writing-style}
\end{table*}

\begin{table*} 
\centering 
\begin{tabular}{cccccc} 
\hline 
metric  & kendall & pearson & spearman & avg-corr & acc \\ \hline
gemini-1.5-pro & 0.3967 & 0.4221 & 0.4226 & 0.4138 & 0.6227 \\ 
bleurt20-mean & 0.3264 & 0.4046 & 0.4160 & 0.3823 & 0.5413 \\ 
gpt-4o & 0.3630 & 0.3836 & 0.3923 & 0.3796 & 0.5787 \\ 
bleurt20-max & 0.3263 & 0.3541 & 0.4157 & 0.3654 & 0.5467 \\ 
bertscore-f1-mean & 0.3021 & 0.3562 & 0.3850 & 0.3478 & 0.5733 \\ 
deepseekv3 & 0.3183 & 0.3434 & 0.3472 & 0.3363 & 0.5867 \\ 
bertscore-f1-max & 0.2883 & 0.3344 & 0.3701 & 0.3309 & 0.5653 \\ 
deltableu-micro & 0.2261 & 0.1678 & 0.2926 & 0.2288 & 0.5147 \\ 
rouge-L & 0.1774 & 0.2056 & 0.2276 & 0.2035 & 0.4853 \\ 
rouge-2 & 0.1283 & 0.1722 & 0.1651 & 0.1552 & 0.4560 \\ 
promethesus7bv2 & 0.1353 & 0.1448 & 0.1497 & 0.1432 & 0.5520 \\ 
rouge-1 & 0.0364 & 0.0996 & 0.0465 & 0.0608 & 0.4560 \\ 
\hline
\end{tabular}
\caption{iiyi-en-overall}
\label{tab:iiyi-en-overall}
\end{table*}

\begin{table*} 
\centering 
\begin{tabular}{cccccc} 
\hline 
metric  & kendall & pearson & spearman & avg-corr & acc \\ \hline
gemini-1.5-pro & 0.4339 & 0.4043 & 0.4551 & 0.4311 & 0.6397 \\ 
bertscore-f1-max & 0.3750 & 0.3880 & 0.4755 & 0.4128 & 0.2948 \\ 
bertscore-f1-mean & 0.3579 & 0.3801 & 0.4551 & 0.3977 & 0.2948 \\ 
gpt-4o & 0.3833 & 0.3860 & 0.4016 & 0.3903 & 0.6463 \\ 
deepseekv3 & 0.3599 & 0.3599 & 0.3772 & 0.3657 & 0.6245 \\ 
promethesus7bv2 & 0.2622 & 0.2430 & 0.2769 & 0.2607 & 0.6135 \\ 
rouge-2 & 0.2321 & 0.2496 & 0.2969 & 0.2595 & 0.2904 \\ 
deltableu-micro & 0.1712 & 0.2296 & 0.2200 & 0.2069 & 0.2686 \\ 
rouge-L & 0.1552 & 0.1271 & 0.1990 & 0.1604 & 0.2511 \\ 
rouge-1 & 0.1282 & 0.1144 & 0.1623 & 0.1350 & 0.2467 \\ 
bleurt20-mean & 0.0768 & 0.0829 & 0.0965 & 0.0854 & 0.1987 \\ 
bleurt20-max & 0.0694 & 0.0834 & 0.0879 & 0.0802 & 0.1943 \\ 
\hline
\end{tabular}
\caption{liveqa-en-overall}
\label{tab:liveqa-en-overall}
\end{table*}

\begin{table*} 
\centering 
\begin{tabular}{cccccc} 
\hline 
metric  & kendall & pearson & spearman & avg-corr & acc \\ \hline
gpt-4o & 0.3769 & 0.4124 & 0.3897 & 0.3930 & 0.5412 \\ 
deepseekv3 & 0.3450 & 0.3863 & 0.3594 & 0.3635 & 0.5448 \\ 
gemini-1.5-pro & 0.2921 & 0.3346 & 0.3012 & 0.3093 & 0.5054 \\ 
bleurt20-mean & 0.1898 & 0.2152 & 0.2400 & 0.2150 & 0.2545 \\ 
bleurt20-max & 0.1789 & 0.1912 & 0.2263 & 0.1988 & 0.2616 \\ 
rouge-2 & 0.1307 & 0.1627 & 0.1665 & 0.1533 & 0.2796 \\ 
bertscore-f1-mean & 0.0986 & 0.1420 & 0.1231 & 0.1212 & 0.2867 \\ 
bertscore-f1-max & 0.0983 & 0.1306 & 0.1228 & 0.1172 & 0.2760 \\ 
deltableu-micro & 0.0996 & 0.1119 & 0.1270 & 0.1128 & 0.2832 \\ 
promethesus7bv2 & 0.0657 & 0.0983 & 0.0708 & 0.0783 & 0.3835 \\ 
rouge-1 & 0.0460 & 0.0675 & 0.0583 & 0.0572 & 0.2545 \\ 
rouge-L & -0.0117 & 0.0173 & -0.0139 & -0.0028 & 0.2437 \\ 
\hline
\end{tabular}
\caption{woundcare-zh-factual-consistency-wgold}
\label{tab:woundcare-zh-factual-consistency-wgold}
\end{table*}

\begin{table*} 
\centering 
\begin{tabular}{cccccc} 
\hline 
metric  & kendall & pearson & spearman & avg-corr & acc \\ \hline
deepseekv3 & 0.0863 & 0.0737 & 0.0874 & 0.0825 & 0.6165 \\ 
gemini-1.5-pro & 0.0769 & 0.0712 & 0.0775 & 0.0752 & 0.5771 \\ 
gpt-4o & 0.0496 & 0.0464 & 0.0500 & 0.0487 & 0.5161 \\ 
deltableu-micro & 0.0268 & 0.0335 & 0.0328 & 0.0310 & 0.0108 \\ 
bleurt20-max & 0.0219 & 0.0293 & 0.0268 & 0.0260 & 0.0108 \\ 
promethesus7bv2 & 0.0234 & 0.0302 & 0.0243 & 0.0260 & 0.4946 \\ 
rouge-2 & 0.0060 & 0.0019 & 0.0073 & 0.0051 & 0.0108 \\ 
rouge-1 & -0.0074 & -0.0154 & -0.0091 & -0.0106 & 0.0108 \\ 
bertscore-f1-max & -0.0141 & -0.0032 & -0.0173 & -0.0115 & 0.0143 \\ 
bleurt20-mean & -0.0127 & -0.0067 & -0.0155 & -0.0116 & 0.0108 \\ 
rouge-L & -0.0113 & -0.0217 & -0.0138 & -0.0156 & 0.0179 \\ 
bertscore-f1-mean & -0.0176 & -0.0333 & -0.0216 & -0.0242 & 0.0108 \\ 
\hline
\end{tabular}
\caption{woundcare-zh-writing-style}
\label{tab:woundcare-zh-writing-style}
\end{table*}

\begin{table*} 
\centering 
\begin{tabular}{cccccc} 
\hline 
metric  & kendall & pearson & spearman & avg-corr & acc \\ \hline
bertscore-f1-max & 0.3379 & 0.4071 & 0.4234 & 0.3895 & 0.2633 \\ 
gemini-1.5-pro & 0.3669 & 0.3975 & 0.3847 & 0.3830 & 0.6200 \\ 
deepseekv3 & 0.3648 & 0.3910 & 0.3892 & 0.3817 & 0.5533 \\ 
rouge-2 & 0.3271 & 0.3979 & 0.4093 & 0.3781 & 0.2700 \\ 
gpt-4o & 0.3600 & 0.3859 & 0.3844 & 0.3767 & 0.5367 \\ 
bertscore-f1-mean & 0.2951 & 0.3633 & 0.3701 & 0.3428 & 0.2700 \\ 
deltableu-micro & 0.2806 & 0.3029 & 0.3534 & 0.3123 & 0.2567 \\ 
bleurt20-max & 0.2483 & 0.3065 & 0.3115 & 0.2888 & 0.2600 \\ 
bleurt20-mean & 0.2108 & 0.2822 & 0.2656 & 0.2529 & 0.2567 \\ 
rouge-1 & 0.1445 & 0.1862 & 0.1825 & 0.1711 & 0.2467 \\ 
promethesus7bv2 & 0.1411 & 0.1223 & 0.1493 & 0.1376 & 0.4800 \\ 
rouge-L & 0.0946 & 0.1466 & 0.1197 & 0.1203 & 0.2100 \\ 
\hline
\end{tabular}
\caption{iiyi-zh-factual-consistency-wgold}
\label{tab:iiyi-zh-factual-consistency-wgold}
\end{table*}

\begin{table*} 
\centering 
\begin{tabular}{cccccc} 
\hline 
metric  & kendall & pearson & spearman & avg-corr & acc \\ \hline
gemini-1.5-pro & 0.1214 & 0.1227 & 0.1248 & 0.1230 & 0.6000 \\ 
gpt-4o & 0.1002 & 0.1041 & 0.1045 & 0.1029 & 0.5033 \\ 
deltableu-micro & 0.0818 & 0.1025 & 0.1000 & 0.0948 & 0.0500 \\ 
deepseekv3 & 0.0643 & 0.0621 & 0.0670 & 0.0645 & 0.5000 \\ 
bertscore-f1-mean & 0.0249 & 0.0407 & 0.0305 & 0.0320 & 0.0467 \\ 
bertscore-f1-max & 0.0227 & 0.0347 & 0.0278 & 0.0284 & 0.0400 \\ 
bleurt20-max & 0.0183 & 0.0175 & 0.0223 & 0.0194 & 0.0267 \\ 
rouge-2 & -0.0105 & 0.0081 & -0.0128 & -0.0050 & 0.0200 \\ 
rouge-1 & -0.0162 & -0.0245 & -0.0198 & -0.0201 & 0.0333 \\ 
promethesus7bv2 & -0.0173 & -0.0267 & -0.0178 & -0.0206 & 0.4833 \\ 
bleurt20-mean & -0.0239 & -0.0241 & -0.0292 & -0.0258 & 0.0333 \\ 
rouge-L & -0.0323 & -0.0306 & -0.0394 & -0.0341 & 0.0433 \\ 
\hline
\end{tabular}
\caption{iiyi-zh-writing-style}
\label{tab:iiyi-zh-writing-style}
\end{table*}

\begin{table*} 
\centering 
\begin{tabular}{cccccc} 
\hline 
metric  & kendall & pearson & spearman & avg-corr & acc \\ \hline
bertscore-f1-max & 0.3106 & 0.3802 & 0.3902 & 0.3604 & 0.2304 \\ 
bertscore-f1-mean & 0.3106 & 0.3802 & 0.3902 & 0.3604 & 0.2304 \\ 
bleurt20-max & 0.2720 & 0.3210 & 0.3408 & 0.3112 & 0.1918 \\ 
bleurt20-mean & 0.2720 & 0.3210 & 0.3408 & 0.3112 & 0.1918 \\ 
gemini-1.5-pro & 0.2982 & 0.3014 & 0.3086 & 0.3028 & 0.5792 \\ 
deepseekv3 & 0.2871 & 0.2973 & 0.2973 & 0.2939 & 0.5740 \\ 
gpt-4o & 0.2830 & 0.3014 & 0.2937 & 0.2927 & 0.5302 \\ 
rouge-2 & 0.2503 & 0.3112 & 0.3145 & 0.2920 & 0.2162 \\ 
deltableu-micro & 0.2406 & 0.2454 & 0.3036 & 0.2632 & 0.2111 \\ 
rouge-1 & 0.1472 & 0.1751 & 0.1860 & 0.1694 & 0.2033 \\ 
rouge-L & 0.1336 & 0.1593 & 0.1695 & 0.1542 & 0.1853 \\ 
promethesus7bv2 & 0.1251 & 0.1596 & 0.1305 & 0.1384 & 0.4659 \\ 
\hline
\end{tabular}
\caption{meddialog-zh-factual-consistency-wgold}
\label{tab:meddialog-zh-factual-consistency-wgold}
\end{table*}

\begin{table*} 
\centering 
\begin{tabular}{cccccc} 
\hline 
metric  & kendall & pearson & spearman & avg-corr & acc \\ \hline
rouge-L & 0.0880 & 0.1321 & 0.1080 & 0.1094 & 0.0309 \\ 
rouge-1 & 0.0841 & 0.1218 & 0.1032 & 0.1030 & 0.0347 \\ 
bertscore-f1-max & 0.0744 & 0.1168 & 0.0914 & 0.0942 & 0.0232 \\ 
bertscore-f1-mean & 0.0744 & 0.1168 & 0.0914 & 0.0942 & 0.0232 \\ 
rouge-2 & 0.0700 & 0.0912 & 0.0859 & 0.0823 & 0.0219 \\ 
deltableu-micro & 0.0772 & 0.0433 & 0.0950 & 0.0718 & 0.0270 \\ 
gemini-1.5-pro & 0.0623 & 0.0664 & 0.0625 & 0.0637 & 0.7169 \\ 
bleurt20-max & 0.0403 & 0.0937 & 0.0494 & 0.0611 & 0.0193 \\ 
bleurt20-mean & 0.0403 & 0.0937 & 0.0494 & 0.0611 & 0.0193 \\ 
deepseekv3 & 0.0529 & 0.0592 & 0.0532 & 0.0551 & 0.6718 \\ 
gpt-4o & 0.0106 & 0.0048 & 0.0106 & 0.0087 & 0.5985 \\ 
promethesus7bv2 & 0.0163 & -0.0094 & 0.0166 & 0.0078 & 0.5611 \\ 
\hline
\end{tabular}
\caption{meddialog-zh-writing-style}
\label{tab:meddialog-zh-writing-style}
\end{table*}

\end{document}